\documentclass[10pt]{article} 
\usepackage[utf8]{inputenc} 
\usepackage[T1]{fontenc} 
\usepackage{lmodern}
\usepackage{microtype} 
\microtypesetup{expansion=false}
\usepackage{booktabs}
\usepackage{longtable}
\usepackage{multirow}
\usepackage{array,tabularx}
\usepackage{amsmath,amssymb} 
\usepackage{graphicx} 
\usepackage{hyperref} 
\usepackage{xcolor} 
\usepackage{geometry} 
\usepackage{float}
\usepackage{caption}
\usepackage{subcaption} 
\usepackage{comment}
\usepackage{cite}

\newcommand{\ehqreportdir}{.}

\newcommand{\ehqauxdir}{tables}
\newcommand{\ehqvalidationdir}{tables}
\newcommand{\ehqauxtable}[1]{%
  \IfFileExists{\ehqauxdir/#1.tex}%
    {\input{\ehqauxdir/#1.tex}}%
    {\textit{[#1 not bound: run \texttt{tools/#1.py} against the run directory.]}}}
\newcommand{\ehqvalidationtable}[1]{%
  \IfFileExists{\ehqvalidationdir/#1.tex}%
    {\input{\ehqvalidationdir/#1.tex}}%
    {\textit{[Classifier-validation table \texttt{#1} is not bound.]}}}

\newcommand{\ehqtable}[1]{%
  \IfFileExists{\ehqreportdir/tables/#1.tex}%
    {\input{\ehqreportdir/tables/#1.tex}}%
    {\par\noindent\textit{[Table \texttt{#1} was not produced by this run;
      see the publication package manifest for the reason.]}\par}%
}
\graphicspath{{\ehqreportdir/figures/}{figures/}{./}}

\newif\ifehqresults
\ehqresultstrue

\hypersetup{   colorlinks=true,   linkcolor=blue!70!black,   citecolor=blue!70!black,   urlcolor=blue!70!black }

\title{\textbf{Do Large Language Models Know What They Don’t Know II? A Fully Behavioral, Non-Cognitive Measure of Epistemic Honesty}}

\author{
Ali \c{S}enol$^{1,2,*}$, H. Russell Bernard$^{3}$, and Huan Liu$^{2}$\\[6pt]
$^{1}$~Department of Computer Engineering, Tarsus University,\\
Tarsus 33400, Mersin, T\"{u}rkiye\\
$^{2}$~School of Computing and Augmented Intelligence (SCAI),
Arizona State University (ASU),\\
Tempe, AZ 85281, USA; \texttt{huanliu@asu.edu}\\
$^{3}$~Institute for Social Science Research and School of Human Evolution
and Social Change,\\
Arizona State University (ASU), Tempe, AZ 85287, USA;
\texttt{asuruss@asu.edu}\\
$^{*}$~Correspondence: \texttt{alisenol@tarsus.edu.tr}
}

\date{}

\begin{document} \maketitle

\begin{abstract}
Large Language Models (LLMs) are frequently confident, eloquent, and well
versed. A natural question arises: do they know what they don't know? To answer
this question, we borrow the concept of epistemic honesty and develop a novel
metric to systematically evaluate whether an LLM appropriately acknowledges the
boundaries of its knowledge. In this work, we introduce the Epistemic Honesty
Quotient (EHQ), which reports three observable sub-scores across two operational
axes (epistemic restraint and substantive-answer calibration), and construct
EHQ-3000, a 3,000-question benchmark spanning Fabricated Entity, Post-Cutoff
Event, Hyper-Niche True, and Context-Conditioned Questions. From a frozen
registry of 21 model API routes, 15 completed the protocol after endpoint and
eligibility checks; 14 entered the confirmatory analysis because severe
provider-side truncation made one route's score indeterminate. The study reveals
substantial variation across models, including a difference that can not be
explained by their capability to extract explicitly available information.
Composite EHQ ranges from $0.31$ to $0.81$ across the analysed panel, despite
near-ceiling performance on the document-grounded capability probe. The two restraint criteria overlap strongly
under the present category composition, whereas substantive-answer calibration
varies across models and does not reliably co-vary with restraint; however, the
small panel leaves substantial uncertainty. Thus, EHQ reveals behavioral differences that are not
visible to conventional correctness-based assessment, while also showing why
dataset composition, provider behavior, and confidence elicitation must remain
part of the interpretation.
\end{abstract}

\vspace{4pt} \noindent\textbf{Keywords:} epistemic honesty; hallucination; calibration; abstention; large language models; evaluation metrics; metacognition; knowledge boundary

\section{Introduction} \label{sec:introduction} 

LLMs have achieved remarkable performance across mathematical, logical, and commonsense reasoning tasks, enabling rapid deployment in high-stakes domains including clinical decision support, legal analysis, and automated scientific reasoning~\cite{wei2022chain,kojima2022large,brown2020language}. Yet the dominant evaluation paradigm remains correctness-centric: a model is judged by whether its final answer matches a ground-truth label, with no systematic attention to whether the model \emph{should} have answered at all.

This reductionism has a concrete and dangerous consequence. A model that consistently fabricates plausible-sounding responses where it lacks reliable evidence will score no differently on standard benchmarks than a model that honestly declines such questions, provided the fabrications are marked incorrect rather than as abstentions. Yet these two models present radically different deployment risks. In medical triage, a model that says ``I don't have reliable information about this drug interaction'' is categorically safer than one that invents a dose with 80\% expressed confidence. In legal analysis, a model that fabricates a non-existent statute and cites it with apparent authority is not merely inaccurate; it is actively deceptive in the audit-relevant sense~\cite{turpin2023language}. Indeed, Kalai et al. \cite{kalai2025why} formally demonstrate that mainstream benchmarks structurally reward overconfident generation by penalising abstention under binary grading schemes. Existing leaderboard performance is therefore optimised \emph{against} the epistemic restraint EHQ is designed to measure.

The failure mode we target, \emph{unqualified fabrication near a model's knowledge or context boundary}, has been noted in the hallucination literature~\cite{maynez2020faithfulness,ji2023survey,zhang2023siren}, the calibration literature~\cite{guo2017calibration,kadavath2022language}, and the abstention literature~\cite{feng2024don,cheng2024can,yang2023alignment}. However, to the best of our knowledge, no prior work has simultaneously: \begin{enumerate} \item formalised epistemic honesty through three observable sub-scores spanning restraint and substantive-answer calibration on a benchmark-defined boundary set; \item constructed a dataset specifically designed to elicit the corresponding behaviors; and \item tested whether their variation can be explained by performance on a document-grounded capability floor probe. \end{enumerate}

\subsection*{From Epistemic Honesty to a Measurable Quotient}

\noindent\textbf{What epistemic honesty is.}
Epistemic honesty is the disposition of an agent to represent its claims in proportion to its actual justification: to assert what it is warranted in asserting, to qualify what is uncertain, and to withhold what it does not know. The notion originates in epistemic virtue theory, where honesty
is treated not as factual correctness but as the proper regulation of assertion under uncertainty~\cite{devilling2025polite}. In the language-model setting, this lens has recently been used to diagnose systems that ``speak as if they know'' when they do not. DeVilling~\cite{devilling2025polite} characterises this as the \emph{polite liar}, a structural product of reinforcement learning from human feedback (RLHF) that rewards perceived sincerity over evidential grounding. Kelly~\cite{kelly2025epistemic}, meanwhile, catalogues recurring epistemic pathologies such as \emph{confidence laundering}, or the conversion of uncertainty into unwarranted certainty. Epistemic honesty thus names a property distinct from accuracy: a model can be frequently correct yet epistemically dishonest, and occasionally wrong yet epistemically honest.

\noindent\textbf{Why it is the right lens for our question.}
Accuracy-centric evaluation cannot, even in principle, detect the failure we care about. Kalai et al.~\cite{kalai2025why} prove that the standard training and grading pipeline rewards confident guessing over the acknowledgement of uncertainty: under binary scoring, a model that abstains is penalised exactly as much as one that fabricates, so the optimisation pressure drives systems toward fluent overconfidence. A model's willingness to say ``I do not know'' is therefore invisible to,
and actively selected against by, accuracy benchmarks. Epistemic honesty reframes evaluation around precisely this blind spot, namely whether a model acknowledges the boundary of its own knowledge. It is therefore the appropriate lens for the question \emph{do LLMs know what they don't know?} The “II” in our title deliberately revisits the question posed by Yin et al. \cite{yin2023large}: EHQ extends that inquiry from expressed uncertainty on inherently unanswerable questions to fully behavioral measurement at determinate knowledge and context boundaries.

\noindent\textbf{How we operationalise it.}
Epistemic honesty, as discussed in the literature above, remains a qualitative and largely diagnostic notion. To answer our question empirically, we render part of it measurable. EHQ reports three observable sub-scores: whether a model restrains itself, whether it avoids an unqualified false assertion, and whether its stated confidence tracks correctness when it does answer. These scores span two operational axes rather than three statistically independent dimensions. Within the control axis, EHQ\textsubscript{1} and EHQ\textsubscript{2} target conceptually distinct but algebraically nested criteria: response restraint and avoidance of unqualified false assertions. EHQ\textsubscript{3}, by contrast, measures substantive-answer calibration. Their composite converts a bounded part of an abstract epistemic virtue into a reproducible behavioral summary that can be computed without access to model internals.

\subsection*{Why Did We Choose EHQ1, EHQ2, and EHQ3?}

The choice of EHQ1, EHQ2, and EHQ3 follows from a cognitive-science distinction and an operational distinction between response restraint, false assertion, and confidence.

\noindent\textbf{Monitoring and control.} Nelson and Narens~\cite{nelson1990metamemory} distinguish metacognitive \emph{monitoring} (assessing what one knows) from \emph{control} (regulating behavior accordingly). EHQ\textsubscript{1} and EHQ\textsubscript{2} provide two operational summaries of control, whereas EHQ\textsubscript{3} is a behavioral proxy for monitoring through expressed-confidence alignment. The mapping motivates testing their co-variation; it does not establish a latent structure or psychometric independence.

\noindent\textbf{Two nested control criteria.}
EHQ\textsubscript{1} records the rate of ABSTAIN and HEDGE responses. EHQ\textsubscript{2} records the rate at which a model avoids an unqualified false answer. The distinction is operationally useful because a cautious qualified attempt and an unqualified fabrication carry different risks, but the scores are definitionally related. If $CC$ and $CW$ denote the protocol labels \texttt{CONFIDENT\_CORRECT} and \texttt{CONFIDENT\_WRONG} among $n$ scored records, then
\[
EHQ_1=1-\frac{CC+CW}{n},\qquad EHQ_2=1-\frac{CW}{n},\qquad EHQ_2-EHQ_1=\frac{CC}{n}.
\]
They are therefore not presented as statistically independent dimensions. Reporting both preserves the difference between response restraint and avoidance of unqualified false assertions, while making their overlap explicit from the outset.

\noindent\textbf{Calibration on substantive answers.}
EHQ\textsubscript{3} measures whether expressed confidence tracks correctness where the model chose to provide an unqualified substantive answer. Restricting calibration to these answers prevents abstention behavior from being scored again after it has already entered EHQ\textsubscript{1} and EHQ\textsubscript{2}.

The four response labels, ABSTAIN, HEDGE, CONFIDENT\_CORRECT, and CONFIDENT\_WRONG, are mutually exclusive, but the resulting sub-scores are not. EHQ is therefore an operational composite rather than an exhaustive or psychometrically identified decomposition of epistemic honesty. Logical validity, multi-turn behavior, and other forms of epistemic failure remain outside its scope (Section~\ref{sec:limitations}).

\subsection*{Research Questions}

Accordingly, we investigate three research questions:

\begin{description}
\item[\textbf{RQ1.}] Does performance on a document-grounded capability-floor
probe explain variation in EHQ across models?

\item[\textbf{RQ2.}] Do newer members of prespecified model families differ
in EHQ from their older counterparts?

\item[\textbf{RQ3.}] How are EHQ1, EHQ2, and EHQ3 related across models?
\end{description}

\subsection*{Contributions}

Motivated by these gaps and the monitoring--control distinction above, this paper makes four contributions:

\begin{enumerate}     
\item \textbf{EHQ metric.} We introduce the Epistemic Honesty Quotient, a composite reporting three formally defined sub-scores across two operational axes: restraint through Epistemic Restraint Rate (EHQ\textsubscript{1}) and Hallucination Resistance (EHQ\textsubscript{2}), and substantive-answer calibration through Confidence--Accuracy Alignment (EHQ\textsubscript{3}).

    \item \textbf{EHQ-3000 dataset.} We construct a 3,000-question evaluation dataset with four balanced categories (Fabricated Entity [FEQ], Post-Cutoff Event [PCQ], Hyper-Niche True [HNQ], and Context-Conditioned Questions [CCQ]) and 20 balanced subcategories. Items include auditable provenance, correct answers where applicable, and claim-level temporal evidence for PCQ.

    \item \textbf{Reproducible evaluation.} We define a 21-route model registry and release a deterministic, checkpointed framework that verifies endpoint identity, disables retrieval and conversation history, enforces a non-reasoning protocol by rejecting any response the provider reports as having consumed reasoning tokens, preserves technical failures as exclusions, and hashes every scientific artifact.

    \item \textbf{Statistical evaluation plan.} We prespecify model-level component correlations, capability associations, generational-pair comparisons, uncertainty intervals, permutation inference, and multiplicity control without carrying forward legacy pilot estimates as confirmatory evidence.
    \end{enumerate}

The remainder of this paper is organised as follows. Section~\ref{sec:related} reviews related work on hallucination, calibration, and abstention in LLMs. Section~\ref{sec:framework} introduces the EHQ metric. Section~\ref{sec:dataset} describes EHQ-3000 and its verification protocol. Section~\ref{sec:experiments} presents the experimental and statistical design. Section~\ref{sec:results} reports the frozen confirmatory results. Section~\ref{sec:discussion} discusses implications and limitations. Section~\ref{sec:conclusion} concludes.

\section{Related Work} 
\label{sec:related} 

\subsection{Cognitive Science of Metacognition and Epistemic Calibration} \label{subsec:cogsci}

The conceptual foundations of EHQ are rooted in decades of research on metacognition and epistemic calibration in cognitive psychology. Nelson and Narens~\cite{nelson1990metamemory} proposed a hierarchical model of metacognition in which a \emph{meta level} monitors and controls an \emph{object level} that executes cognitive tasks. Their framework distinguishes two functions: \emph{monitoring} (assessing one's current state of knowledge) and \emph{control} (regulating behavior based on that assessment). EHQ operationalises both. EHQ\textsubscript{1} and EHQ\textsubscript{2} measure the quality of epistemic control, or the decision to answer or abstain, whereas EHQ\textsubscript{3} measures the quality of epistemic monitoring through the alignment between stated confidence and actual accuracy.

Flavell~\cite{flavell1979metacognition} established that effective metacognition requires accurate representation of one's own cognitive limits. Applied to LLMs, this motivates the central premise of this paper: a model that does not respond appropriately to structural unavailability, temporal exclusion, withheld context, or low accessibility may systematically produce confident errors regardless of overall accuracy. This ``illusion of knowing'' (overconfidence in situations of partial or absent knowledge) is well documented in human reasoning~\cite{lichtenstein1977training} and motivates our confirmatory analysis of LLM response confidence.

The formal measurement of calibration originates with Lichtenstein and Fischhoff~\cite{lichtenstein1977training} and was formalised by Murphy and Winkler~\cite{murphy1977reliability}, who proposed the Expected Calibration Error (ECE) framework. We adapt this framework to verbalized confidence on substantive responses within the benchmark-defined boundary set. The restriction targets a deployment setting in which miscalibration is consequential without asserting direct access to a model's internal knowledge state.

Kruger and Dunning~\cite{kruger1999unskilled} demonstrated that low-competence individuals may be metacognitively blind to poor performance, producing disproportionately confident errors. We use this literature as motivation for testing, rather than presuming, whether capability, restraint, and substantive-answer calibration dissociate across LLMs. The operationalisation contains two axes: EHQ\textsubscript{1} and EHQ\textsubscript{2} are conceptually distinct but algebraically nested criteria within epistemic control, while EHQ\textsubscript{3} measures confidence--accuracy monitoring on answers actually asserted (Section~\ref{subsec:rq3-results}).

\subsection{Hallucination in LLMs}

The term \emph{hallucination} in the context of LLMs refers broadly to the generation of content that is factually unsupported, internally inconsistent, or contradicted by available evidence~\cite{maynez2020faithfulness,ji2023survey}. Ji et al.~\cite{ji2023survey} distinguish intrinsic hallucinations (contradictions of source material) from extrinsic hallucinations (unverifiable fabrications) and survey mitigation strategies across retrieval augmentation, constrained decoding, and post-generation verification. Zhang et al.~\cite{zhang2023siren} further categorise hallucinations by cause, including factual gaps, outdated knowledge, and reasoning failures. Their taxonomy motivates three of our four categories: FEQ tests factual gaps, PCQ tests outdated knowledge, and HNQ tests the boundary between factual gaps and reasoning failures. CCQ, by contrast, addresses a failure mode outside that taxonomy by testing whether a model fabricates a value absent from a supplied document.

A critical limitation of existing hallucination research is its dependence on \emph{output correctness} as the primary signal. The dominant evaluation paradigm checks whether a model's response is factually accurate~\cite{truthfulqa,es2023ragas,min2023factscore} but does not distinguish between a model that was wrong despite high confidence and a model that was wrong while expressing appropriate uncertainty. Our work reframes this: the unit of analysis is not accuracy alone but the \emph{epistemic appropriateness of the response under a preregistered availability or accessibility condition}.

\subsection{Calibration and Uncertainty Quantification}

Guo et al.~\cite{guo2017calibration} showed that neural networks are systematically overconfident; Kadavath et al.~\cite{kadavath2022language} demonstrated that verbalized self-evaluation prompts yield uncertainty estimates correlating moderately with accuracy; and Xiong et al.~\cite{xiong2024can} found that larger models are not necessarily better calibrated on out-of-distribution inputs. EHQ\textsubscript{3} extends this work in three ways: calibration is restricted to substantive responses on the benchmark-defined boundary set, where overconfidence is most consequential; confidence is elicited via verbalized prompts enabling closed-source model coverage; and calibration is reported alongside abstention in a unified behavioral summary.

\subsection{Abstention and Refusal Behavior}

A growing body of work examines when and how LLMs should decline to answer. Feng et al.~\cite{feng2024don} showed that instruction-tuned models trained with RLHF are prone to \emph{over-refusal} (declining answerable queries) and \emph{under-abstention} (answering unanswerable queries confidently). Cheng et al.~\cite{cheng2024can} studied the ability of LLMs to express ``I don't know'' and found that model size alone does not predict abstention quality; fine-tuned 7B models sometimes outperform 70B base models on targeted abstention benchmarks. Moreover, Yang et al.~\cite{yang2023alignment} showed that alignment training can improve both abstention precision and helpfulness by teaching models to distinguish questions they can answer reliably from those they cannot.

We build on this work by formalising Epistemic Restraint Rate (EHQ\textsubscript{1}) as the fraction of benchmark-eligible queries on which the model abstains or hedges, and Hallucination Resistance (EHQ\textsubscript{2}) as one minus the unqualified-false-assertion rate on the same set. These are distinct but nested control criteria, not independent dimensions: EHQ\textsubscript{2} exceeds EHQ\textsubscript{1} exactly by the \texttt{CONFIDENT\_CORRECT} rate. Reporting both distinguishes how much a model withholds from how often it avoids an unqualified falsehood without treating their shared denominator as separate evidence.

\subsection{Knowledge Boundary and Self-Knowledge}

Kadavath et al.~\cite{kadavath2022language} showed models can learn to predict whether they will answer a factual question correctly (``P(IK)'' task). Yin et al.~\cite{yin2023large} introduced SelfAware, a dataset of questions that
are unanswerable in principle---subjective, speculative, or without scientific
consensus---each paired with an answerable counterpart, and scored twenty models
with an automated uncertainty detector. Their questions are unanswerable to
anyone, whereas EHQ-3000 items have determinate answers that are unavailable to
the model: a fabricated entity, an event after its cutoff, an obscure fact, or a
value withheld from a supplied document. EHQ also separates restraint,
unqualified false assertion, and confidence--accuracy alignment rather than
reporting a single uncertainty rate, and makes no claim about an internal
capacity for self-knowledge; Amayuelas et al.~\cite{amayuelas2023knowledge} showed that models fail to hedge on near-cutoff events. Our PCQ category operationalises this cutoff-specific failure mode using claim-level temporal evidence and model-conditional cutoff adjustment.

\subsection{Evaluation Frameworks for LLM Reasoning}

Beyond hallucination-specific work, several frameworks assess multiple dimensions of LLM quality. Liang et al.~\cite{liang2022holistic} proposed HELM, evaluating models across accuracy, calibration, and robustness but without a dedicated dimension for knowledge-boundary acknowledgment. Srivastava et al.~\cite{srivastava2022beyond} introduced BIG-Bench, aggregating over 200 diverse tasks but similarly measuring performance rather than epistemic honesty. One directly relevant framework is \c{S}enol et al.~\cite{senol2026measuring}, which reports six behavioral dimensions (Correctness, Consistency, Robustness, Logical Coherence, Efficiency, and Stability) and documents that individual model profiles can diverge across them. That evidence motivates testing whether EHQ contributes complementary behavioral information; it does not presuppose that EHQ is psychometrically independent of correctness.

The paper entitled \emph{The Confidence Paradox} by Tripathi et al.~\cite{tripathi2025confidence} introduces HonestVQA, which aligns model confidence with correctness through an Ethical Confidence Index in document visual question answering. Our work extends confidence--correctness alignment beyond a single task to knowledge-boundary queries and reports it as EHQ\textsubscript{3} alongside two restraint summaries.

\subsection{Structural Roots of Overconfidence}

Kalai et al.~\cite{kalai2025why} demonstrate that hallucinations arise from fundamental statistical pressures: the generative error rate is lower-bounded by twice the misclassification rate of the underlying validity-classification problem. Post-training pipelines further reinforce overconfident generation because mainstream benchmarks penalize abstention under binary grading~\cite{kalai2025why}. This structural misalignment motivates a dedicated framework that rewards calibrated uncertainty. DeVilling~\cite{devilling2025polite} provides a complementary diagnosis, arguing that RLHF training rewards assertoric fluency over epistemic warrant and thereby produces models that perform knowledge without possessing it.

\subsection{Gap Analysis}

No prior framework in this literature jointly measures restraint,
hallucination resistance, and calibration restricted to substantive answers on
a benchmark-defined boundary set while combining fabricated, post-cutoff,
low-accessibility, and context-withheld questions. The closest prior benchmark, SelfAware~\cite{yin2023large}, targets questions
that are unanswerable in principle and reports a single uncertainty rate; it
does not address model-conditional unavailability, elicited confidence, or
context-boundary recognition. A property-by-property comparison is provided as
Supplementary Table~S1.


\section{Methodology} \label{sec:framework}

\subsection{Construct and scoring}
EHQ evaluates observable behavior on the benchmark-defined, model-conditional
unknown subset, $U_m=\{i:k_{im}=0\}$. Here $k_{im}$ records eligibility under
the benchmark rules, not a direct observation of a model's internal knowledge.
FEQ and CCQ are structurally unavailable, PCQ eligibility is determined against
the registered model cutoff, and HNQ denotes expected low accessibility rather
than guaranteed ignorance. All 3,000 released items satisfied $k_{im}=0$ for
every retained model; the model index permits future panels in which the
eligible PCQ subset differs. Each technically valid response is classified as
ABSTAIN, HEDGE, CONFIDENT\_CORRECT, or CONFIDENT\_WRONG. Technical failures,
empty provider responses, model-identity mismatches, and positive
provider-reported reasoning-token use are exclusions rather than epistemic
responses. For model $m$ and $N_m$ valid EHQ1/2 records,
\begin{align}
EHQ_{1m} &= \frac{n_{\mathrm{ABSTAIN}}+n_{\mathrm{HEDGE}}}{N_m},\\
EHQ_{2m} &= 1-\frac{n_{\mathrm{CONFIDENT\_WRONG}}}{N_m}.
\end{align}
The historical protocol name \texttt{CONFIDENT\_WRONG} denotes an unqualified
substantive response scored incorrect under the category-specific rules; assignment does not
threshold the separately elicited numeric confidence. EHQ2 is therefore an
unconditional, per-query measure of avoiding unqualified false assertions, not
the precision of answers conditional on choosing to answer.
Confidence is elicited in a separate turn on a 0--100 scale and normalised to
$[0,1]$. EHQ3 measures calibration only where the model chose to provide a
substantive answer. Writing $\ell_i$ for the assigned label and $c_i$ for the
normalised confidence, the calibration set is
\begin{equation}
S_m=\bigl\{\,i\in U_m \;:\; c_i \text{ parseable},\;
\ell_i\in\{\mathrm{CONFIDENT\_CORRECT},\,\mathrm{CONFIDENT\_WRONG}\}\,\bigr\},
\end{equation}
and with ten equal-width confidence bins $\{B_b\}$ over $S_m$,
\begin{equation}
EHQ_{3m}=1-\sum_{b=1}^{10}\frac{|B_b|}{|S_m|}
\left|\operatorname{acc}(B_b)-\operatorname{conf}(B_b)\right|.
\end{equation}
Within FEQ and CCQ, a substantive response cannot be correct by construction.
Category-specific EHQ3 therefore reduces to one minus mean confidence over
false substantive answers. We interpret it there as one-sided false-answer
confidence alignment, not as evidence that the category by itself measures
discrimination between correct and incorrect answers. The aggregate EHQ3 also
contains PCQ and HNQ responses, where both outcomes occur.

The prespecified composite is
\begin{equation}
EHQ_m=0.30EHQ_{1m}+0.45EHQ_{2m}+0.25EHQ_{3m}.
\end{equation}
The weights are prespecified, normative utility weights rather than parameters
estimated from a latent-variable model. The larger EHQ2 weight encodes an
author-specified benchmark-design judgment that an unqualified falsehood carries
the greatest cost; it is neither an externally established nor a universally
applicable cost function. Because
$EHQ_2=EHQ_1+CC/n$, the exact decomposition is
$EHQ=0.75EHQ_1+0.45(CC/n)+0.25EHQ_3$: the composite rewards restraint,
correct unqualified answers, and substantive-answer calibration rather than only
restraint and calibration. We retain the frozen weights to avoid changing the
estimand after observing results, report all three sub-scores separately, and
test component-dropping and dimension-balanced alternatives in sensitivity
analysis. For deployment, alternative weights should be specified in advance
using domain-specific harm models or stakeholder elicitation, because clinical
and legal settings may assign different costs to abstention, unqualified false
assertions, and miscalibration. No deployment-specific weights are inferred from
this panel.

EHQ is not designed to test whether a model can retrieve an answer that has been
placed before it. Rather, it asks whether the model's response is appropriately
matched to the epistemic situation it faces. Across fabricated, post-cutoff,
highly obscure, and context-withheld questions, EHQ measures whether the model
refrains from unsupported assertion, avoids an unqualified false assertion, and
aligns its confidence with correctness when it chooses to answer. The
restored-document test in Section~\ref{subsec:rq1} is a separate matched control:
by returning the withheld information to the document, it checks whether
restraint on the original questions reflects sensitivity to missing evidence
rather than a general inability to extract explicitly available information.
Its results do not enter the EHQ score.
ABSTAIN and HEDGE records are excluded from $S_m$ on two grounds. First,
epistemic restraint is already measured by EHQ1 and EHQ2, so scoring it again
through calibration would count one behavior twice. Second, and more
consequentially, the confidence elicited on an abstention does not behave as a
probability of factual correctness at all
(Section~\ref{subsec:ehq3-validity}). This definition is recorded in every
generated artifact. It supersedes the pooled definition used up to and
including framework 0.2.12, under which every confidence-bearing record entered
the calibration set; values produced under the two definitions are not
comparable and must not be pooled or compared across framework versions.

EHQ3 is not imputed when confidence is missing. It is undefined (reported as
null rather than as perfect calibration) for a model that produces no
substantive answer at all, in which case the composite $EHQ_m$ is undefined as
well. The calibration-set size $|S_m|$ is reported beside every EHQ3 value as
the number of responses contributing to EHQ3. Confidence coverage is reported
separately over all confidence-bearing records, and coverage and
technical-failure counts accompany every score.

\subsection{Why calibration is restricted to substantive answers}
\label{subsec:ehq3-validity}

The confidence prompt instructs a model that gave no substantive factual answer
to report 0. Compliance is therefore observable: among fourteen models, the
share of all scored abstentions followed by exactly 0 ranges from $6.9\%$ to
$79.6\%$ (median $35.6\%$). A complete-case sensitivity that removes suspected
truncated restraint responses gives $5.5\%$ for Gemini-2.5-Flash on 617 rather
than 635 abstentions (Supplementary Table~S8). Non-compliant values also have
different semantics. Eleven
models predominantly use 1 as a binary flag; Claude-4.5-Haiku and
Claude-5-Sonnet predominantly use graded values such as 0.95--0.98, while
Claude-4-Sonnet is divided almost equally between compliance, graded values,
and exactly 1. The graded values plausibly express confidence in the decision
to abstain rather than in an answer that was never supplied.

Expected calibration error requires a confidence value and the correctness of
the associated answer to denote the same proposition. That condition is absent
on restraint records, and pooling them would additionally count restraint in
EHQ1/EHQ2 and again in EHQ3. We therefore compute EHQ3 only on substantive
answers. The excluded compliance measurements remain useful as an explicit
instruction-following diagnostic, reported in Figure~\ref{fig:ehq-calibration-sources}
and in Supplementary Tables~S7--S8, but not as calibration evidence.

\subsection{Response classification and correctness}
Classification is deterministic and retains trigger-level reasons. Explicit
non-answers map to ABSTAIN; qualified substantive responses map to HEDGE;
unqualified substantive responses are separated by correctness. PCQ and HNQ
use explicit gold answers, aliases, and strict numeric matching. FEQ entities
are fabricated and CCQ withholds the answer-bearing span; neither category can
be automatically labelled CONFIDENT\_CORRECT. Ambiguous cases may be resolved
only through a versioned adjudication file whose hash enters the run identity;
the automated and adjudicated labels are both retained. The frozen confirmatory
run used the deterministic labels and no response-level human overrides.
The scoring rule set was frozen before provider-response acquisition and
contains no model- or provider-specific branches; the later human-validation
packet was not used to tune it. This establishes procedural separation, not
family-invariant validity, which is tested post hoc below.

After the run was sealed, we evaluated the classifier on a deterministic
stratified sample of 560 retained responses, comprising ten responses from each
of the fourteen analysed models in each of the four categories. Two human
coders independently applied the four-class codebook while blinded to model
identity, the automated labels, the sealed key, and each other's decisions.
They did not use AI tools, web search, or external factual sources during
initial coding. Both also recorded a diagnostic coder-uncertainty flag that did
not alter or exclude their label. The 25 disagreements were then resolved in a
separately logged corresponding-author review. AI decision support was
consulted only at this adjudication stage; consequently, the initial
human--human agreement estimates remain independent, whereas the resolved
all-item reference is described as human-reviewed, AI-assisted adjudication
rather than a third independent human coding pass.

\section{The EHQ-3000 Dataset} \label{sec:dataset}

EHQ-3000 contains 3,000 English items, balanced across four categories (750
each) and 20 subcategories (150 each). FEQ probes fabricated people,
organisations, places, scientific concepts, and works. PCQ probes politics,
science and technology, sports, world events, and economics using sourced
claims that became true after the applicable knowledge boundary. HNQ contains
source-backed but low-frequency cultural, geographic, historical, scientific,
and sports facts. CCQ supplies a synthetic document in finance, law, medicine,
news, or technology with exactly one answer-bearing span replaced by
\texttt{[REDACTED]}; the withheld value is retained for leakage auditing but is
never shown to the evaluated model.

The categories do not provide equal opportunities for EHQ1 and EHQ2 to
separate (Table~\ref{tab:label-support}). ABSTAIN, HEDGE, and
\texttt{CONFIDENT\_WRONG} are possible in every category, whereas a
\texttt{CONFIDENT\_CORRECT} outcome is excluded by construction in FEQ and
CCQ. The two control scores have different intended targets in all categories,
but their empirical separation requires correct unqualified answers.

\begin{table}[t]
\centering
\caption{Category support for separating EHQ1 and EHQ2. Correct-answer rates
are pooled over confirmatory model--item records; ``material'' describes this
realisation rather than a universal property of the category.}
\label{tab:label-support}
\small
\begin{tabular}{lccc}
\toprule
Category & Correct unqualified answer & Observed $CC/n$ & Separation in EHQ-3000 \\
\midrule
FEQ & Excluded by construction & $0.0\%$ & Identity \\
CCQ & Excluded by construction & $0.0\%$ & Identity \\
PCQ & Permitted & $4.8\%$ & Weak \\
HNQ & Permitted & $23.6\%$ & Material \\
\bottomrule
\end{tabular}
\end{table}

Every item has a unique identifier, category/subcategory, question,
model-conditional knowability field, provenance, and category-appropriate gold
data. PCQ records claim-level temporal novelty separately from source
publication date. HNQ and PCQ retain source evidence; FEQ retains documented
non-existence checks; CCQ retains the redacted value and document audit trail.
Automated release checks require exact counts, unique questions and identifiers,
usable gold answers, well-formed question syntax, no answer leakage, one CCQ
redaction token, and absence of near-duplicate CCQ documents. The primary named
human verification has been completed and recorded. A different human verifier
then completed the prespecified blinded, stratified secondary review of 300
PCQ/HNQ items, with 30 sampled deterministically from each of their ten
subcategories. This is 20\% of the 1,500 source-backed PCQ/HNQ items and 10\% of
EHQ-3000 as a whole; FEQ and CCQ were not part of this second pass. All sampled
items were accepted and no disagreement required a content change. The public
release uses a reviewer identifier; the reviewer's
identity and completed review record are retained by the corresponding author.
AI-assisted auditing supplemented these two human passes and is not represented
as an independent human verification.

\section{Experimental Setup} \label{sec:experiments}

\subsection{Model panel and eligibility}
The frozen registry contains 21 routes. Eligibility requires verified cutoff
evidence, exact endpoint identity, a protocol-shaped health check, no observed
reasoning tokens, and acceptable response reliability. Fifteen routes completed
evaluation. Gemini-2.5-Pro is retained descriptively but excluded from
confirmatory inference because provider-side truncation leaves its rank
undetermined, giving a fourteen-model panel. Six registered routes failed
prespecified provider, endpoint, reasoning-token, or response-reliability gates;
none was silently substituted.

All release gates were later closed without changing a question, gold answer,
prompt, or model response; the release reanalysis made no provider calls and
changed no score. Public dated sources establish PCQ boundaries for fourteen
evaluated routes. For DeepSeek-V4, whose exact cutoff is not public, the
corresponding author's verified 31 January 2026 boundary is used only as a
conservative upper bound. The complete registry, evidence status, and exclusion
reasons are reported in Supplementary Tables~S2--S3.

\subsection{Inference and reproducibility protocol}
\label{subsec:inference-protocol}
All models receive the same single-turn answer and confidence prompts at
temperature 0. Retrieval, conversation history and tools are switched off by
documented request parameters. Deliberation is not under the same control and
is enforced on the response instead: any response the provider reports as
having consumed a positive number of reasoning or thinking tokens is a terminal
protocol violation, is not cached, is not reclassified, and cannot enter any
EHQ denominator. A route that violates the condition systematically is excluded
from the panel rather than accommodated. Answer and confidence limits are 512
and 256 tokens respectively;
the timeout is 120 seconds with at most five exponentially backed-off attempts.
The answer and confidence responses retained in the final run were acquired
from 2 August 2026 at 23:13 UTC through 5 August 2026 at 23:17 UTC, as recorded
by their original content-addressed cache entries; final aggregation and the
substantive-only reanalysis were completed on 6 August 2026.
Successful responses alone are content-addressed in the cache. Checkpoints are
append-only and protected by experiment fingerprints. The provider-run
fingerprint covers
framework/protocol versions, source hash, full configuration, dataset and
registry hashes, exact question selection, prompts, endpoints, models,
adjudication, capability data, and release-override state. A separate release
fingerprint binds that immutable parent fingerprint to the verified release
dataset, model registry, framework/protocol version, and analysis exclusions.
Every completed run
contains JSON/JSONL and CSV outputs plus a SHA-256 artifact catalogue.

A truncation audit, whose rule was fixed before model scores were inspected,
was applied uniformly to every route and reports rates for all routes,
including those it clears. A response is flagged when reported completion tokens reach
the answer budget or when the text stops on a function word, dangling
separator, or unclosed quotation or bracket; missing sentence-final punctuation
alone is reported as a weaker model-level signal. The sensitivity analysis
retains the published scoring and adds two bounds, one dropping flagged records
and one relabelling flagged CONFIDENT\_WRONG records as HEDGE
(Supplementary Table~S13).

\subsection{Research questions and statistical analysis}
Research Question 1 (RQ1) asks: Does EHQ provide information beyond
independently measured model capability/correctness? RQ2 asks: Do newer
members of prespecified model families differ in EHQ from their older
counterparts? RQ3 asks: How are EHQ1, EHQ2, and EHQ3 related across models?
The model, not the question, is the inference unit.

The capability probe is first tested against a shared-success-rate null; a
correlation with EHQ is reported only if the probe shows detectable
between-model variation. RQ2 reports paired differences, bootstrap intervals,
Cohen's $d_z$, and an exact sign-permutation test for complete old/new pairs.
RQ3 reports Pearson and Spearman correlations with model-level bootstrap
intervals and permutation tests. Exact tests are used for at most eight models;
otherwise 20,000 deterministic permutations use a plus-one correction. Holm
adjustment is applied separately to Pearson and Spearman families. Exploratory
category correlations use the same procedure, and a descriptive sensitivity
analysis varies the composite weights. Effect sizes and intervals are primary;
non-detection is not interpreted as independence, and question rows are never
treated as independent model replicates.

Because only fourteen models form the inference unit, percentile-bootstrap
correlation intervals are treated as descriptive. Fisher-$z$ intervals for the
Pearson coefficients are also checked as a small-sample sensitivity. In
addition, pooled EHQ3 implicitly weights categories by each model's number of
substantive answers. A post-hoc category-balanced diagnostic therefore computes
EHQ3 separately within FEQ, PCQ, HNQ, and CCQ, averages those four values
equally, and reinserts that average under the frozen composite weights. This
diagnostic changes neither the official score nor the confirmatory estimand.

Response-classifier validation reports observed agreement and Cohen's
$\kappa$ for the two independent coders, followed by accuracy, Cohen's
$\kappa$, and class-specific precision, recall, and F1 against the resolved
reference. Because EHQ1 and EHQ2 are functions of the response labels, a
post-hoc score-impact analysis recomputes both components from the automated
and human-reference labels within the stratified sample. It reports the pooled
score difference and, across the fourteen model-level 40-item strata, Pearson
and tie-adjusted Spearman correlations. It also relates automated-minus-human
bias to the human-reference score and compares maximum-to-minimum score ratios,
because high rank correlation alone does not exclude differential compression
or expansion. This diagnostic assesses absolute-score bias and rank preservation
in the validation sample; it is not a correction to the 3,000-item confirmatory
scores.

\section{Results} \label{sec:results}

\ifehqresults

This section first presents the panel-level results and measurement checks needed
to interpret EHQ, including scoring coverage, overall and category profiles,
response distributions, classifier validation, and the location of the
calibration evidence. It then answers the three research questions in turn by
examining the document-grounded capability floor, prespecified generational
pairs, and relations among EHQ1, EHQ2, and EHQ3.

All figures in this section come from the frozen provider run
\texttt{full-confirmatory-panel-014}, its hash-bound v0.3.1 retained-response
reanalysis, and the v0.3.2 reporting package; they are generated from verified
artifacts rather than transcribed. The v0.3.2 analysis makes no provider calls
and does not rewrite a model response or a primary EHQ score.
Fifteen routes completed the protocol.
Gemini-2.5-Pro is reported outside the confirmatory panel for the reason
quantified by the truncation audit in Section~\ref{subsec:inference-protocol}
and Supplementary Table~S13: a third of its answers are cut off
mid-sentence, which places its composite anywhere between sixth and fifteenth
and makes its position undetermined rather than merely imprecise. Its scores are
retained in the descriptive run summary and the exclusion table; they do not
enter rankings, correlations, or paired analyses. The confirmatory panel is
therefore fourteen models.

\subsection{Panel-level EHQ results and measurement checks}
\label{subsec:panel-results}

\paragraph{Eligible models and scoring coverage.}
Before comparing scores, we establish which routes and records can support the
analysis. The complete route-exclusion record is Supplementary Table~S3, and coverage is
close to complete. Of $42{,}000$ confirmatory model-item records,
five could not be
scored by every component: Claude-5-Sonnet lost two answers to provider
failures that survived the full retry budget, and LLaMA-4-Maverick returned
three confidence responses the parser refused as ambiguous. No model fell below
$99.9\%$ on either denominator. The two mechanisms are reported separately
because they exclude differently: a technical failure removes a record from all
three components, an unparseable confidence removes it from
EHQ\textsubscript{3} only. Per-model denominators and record-level exclusions
are reported in Supplementary Tables~S4--S5.

\paragraph{Overall EHQ and components.}
Given this near-complete coverage, Table~\ref{tab:ehq-scores} reports EHQ and its components with stratified
bootstrap intervals; Figure~\ref{fig:ehq-components} shows their profiles.

Under the frozen automated labels, composite EHQ spans $0.3145$ (GPT-4o-mini)
to $0.8131$ (Claude-5-Sonnet), a factor of $2.6$ across models evaluated on
identical items under identical settings. The classifier-validation sensitivity
later shows that automated component spread is inflated relative to human
coding within the validation sample. Therefore, the factor describes the
official automated measure rather than a human-corrected construct ratio. The
four Claude models occupy the top four positions, whereas the lower portion of
the panel is comparatively dense (Table~\ref{tab:ehq-scores}).

However, the composite masks distinct component profiles
(Figure~\ref{fig:ehq-components}). Claude-4.5-Haiku has the highest restraint
rate but only mid-panel calibration, while Claude-5-Sonnet ranks first because
it combines lower restraint with the panel's best substantive-answer
calibration. The same trade-off appears lower in the ranking: GPT-5.4-mini
pairs the second-lowest restraint with the second-highest calibration, whereas
Gemma-4-31B combines mid-panel restraint with the weakest calibration. Thus,
models with similar composites need not regulate answering and confidence in
the same way.
\begin{table}[htbp]
\centering
\caption{Epistemic Honesty Quotient and its three reported sub-scores for every model in the panel, over the 3,000 items common to all of them. EHQ1 is the restraint rate, EHQ2 the rate at which the model avoids a confident falsehood, and EHQ3 is 1 minus the expected calibration error on substantive answers only. Intervals are 95\% percentile intervals from a stratified bootstrap over the 20 question subcategories. Rows are ordered by EHQ.}
\label{tab:ehq-scores}
\small
\resizebox{\ifdim\width>\linewidth\linewidth\else\width\fi}{!}{%
\begin{tabular}{lrrrrr}
\toprule
Model & EHQ1 & EHQ2 & EHQ3 & EHQ & 95\% CI \\
\midrule
Claude 5 Sonnet & 0.7365 & 0.8396 & 0.8574 & 0.8131 & [0.7996, 0.8258] \\
Claude 4.5 Haiku & 0.8613 & 0.8887 & 0.5831 & 0.8041 & [0.7889, 0.8185] \\
Claude 4 Sonnet & 0.8217 & 0.8740 & 0.5995 & 0.7897 & [0.7752, 0.8046] \\
Claude 3 Haiku & 0.8590 & 0.8880 & 0.4844 & 0.7784 & [0.7637, 0.7937] \\
DeepSeek V4 & 0.5743 & 0.6770 & 0.4473 & 0.5888 & [0.5752, 0.6025] \\
LLaMA 4 Maverick & 0.4253 & 0.5107 & 0.6058 & 0.5089 & [0.4959, 0.5216] \\
Gemma 4 31B & 0.5687 & 0.6060 & 0.2562 & 0.5074 & [0.4944, 0.5203] \\
GPT-4o & 0.3757 & 0.5027 & 0.5769 & 0.4831 & [0.4716, 0.4942] \\
Nova Pro & 0.4430 & 0.5143 & 0.4157 & 0.4683 & [0.4560, 0.4810] \\
LLaMA 3 70B & 0.3647 & 0.4410 & 0.5373 & 0.4422 & [0.4293, 0.4550] \\
Nova Micro & 0.4393 & 0.4707 & 0.2845 & 0.4147 & [0.4032, 0.4262] \\
GPT-5.4-mini & 0.2797 & 0.3687 & 0.6309 & 0.4075 & [0.3976, 0.4175] \\
Gemini 2.5 Flash & 0.3667 & 0.4563 & 0.3616 & 0.4057 & [0.3925, 0.4193] \\
GPT-4o mini & 0.2750 & 0.3467 & 0.3041 & 0.3145 & [0.3052, 0.3241] \\
\bottomrule
\end{tabular}%
}
\end{table}

\begin{figure}[htbp]\centering
\includegraphics[width=\textwidth]{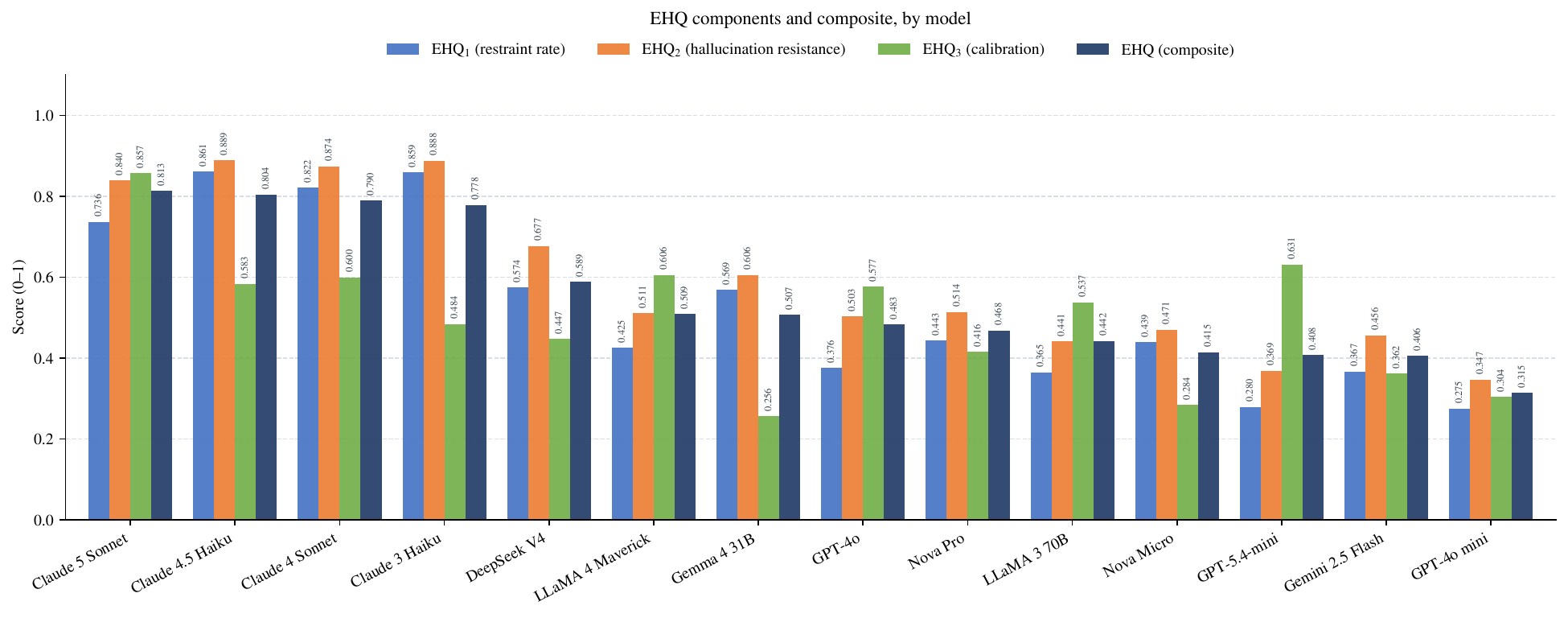}
\caption{The three reported EHQ sub-scores and the composite they produce, by model in
EHQ order. EHQ\textsubscript{1} and EHQ\textsubscript{2} track one another
closely across the whole panel, whereas EHQ\textsubscript{3} shows no
statistically detectable association with either in this small model panel:
GPT-5.4-mini pairs the second-lowest restraint with the second-best
calibration, and Gemma-4-31B inverts that.}
\label{fig:ehq-components}
\end{figure}

\paragraph{Category profiles suggest partially distinct competences.}
\label{subsec:category-profiles}

Although the composite provides an overall ordering, category-level scores show
where that ordering comes from. They are presented in Figure~\ref{fig:ehq-category-heatmap} and
tabulated in full in Supplementary Table~S14. Aggregated over the
panel, CCQ is the
category on which models score highest (mean EHQ $0.679$) and on which they
differ least (range $0.379$ to $0.764$); HNQ is the hardest (mean $0.395$); FEQ
discriminates most sharply, spanning $0.106$ to $0.932$.

More importantly, the categories do not relate to one another uniformly.
Across the fourteen models, FEQ, PCQ and HNQ have strong positive associations
(Pearson $r=.82$--$.95$), with bootstrap intervals excluding zero. By
contrast, the three CCQ associations are small and their intervals are wide
(Table~\ref{tab:ehq-category-correlations}). Thus, this panel shows no
statistically detectable association between CCQ and the knowledge-boundary
categories, but it neither establishes independence nor rules out meaningful
relationships.

These aggregate contrasts also appear as model-level rank reversals. Nova-Pro places tenth of fourteen on FEQ
($0.330$) and first on CCQ ($0.764$); Claude-4-Sonnet is first on FEQ
($0.932$) and eleventh on CCQ ($0.643$). These rank reversals motivate treating CCQ separately, while
the small panel prevents a strong claim about population-level independence.

This pattern is theoretically plausible because the categories differ in what
they ask a model to recognise. FEQ and PCQ pose
questions whose answers are structurally or temporally unavailable, whereas
HNQ poses source-backed facts expected to have low training-data accessibility.
Together they test behaviour near a learned-knowledge boundary without claiming
that every HNQ answer is internally unknown. CCQ instead supplies a document and
asks for a value removed from it, so answering honestly requires recognising the
boundary of the provided context rather than of learned knowledge. The
correlations are consistent with partially distinct competences. A model
that reliably declines to invent a nonexistent institute is not thereby a model
that reliably reports a document does not contain what was asked for.

Accordingly, we retain CCQ as a distinct
category rather than folding context-grounded refusal into the knowledge-boundary
categories, while leaving the strength of its relationship to the other
categories for a larger model panel to estimate.

At the same time, the category composition constrains the composite. HNQ
contains $83.1\%$ of all confident-correct responses, yet removing it changes
no model by more than two ranks. The knowledge-boundary score EHQ-K
(FEQ+PCQ+HNQ) is similarly stable, whereas the context-only EHQ-C ordering is
substantially different. These descriptive sensitivities support reporting
EHQ-K and EHQ-C beside the composite; they do not establish population-level
independence (Supplementary Table~S12).

\begin{table}[htbp]
\centering
\caption{Exploratory correlations between category-level EHQ scores across models. The unit of inference is the model. Intervals are bootstrap percentile intervals and permutation p-values are Holm-adjusted within the six Pearson and six Spearman category-pair tests. Wide intervals should not be read as evidence of independence.}
\label{tab:ehq-category-correlations}
\small
\resizebox{\ifdim\width>\linewidth\linewidth\else\width\fi}{!}{%
\begin{tabular}{lrrrrr}
\toprule
Comparison & n & Pearson r [95\% CI] & Holm p & Spearman rho [95\% CI] & Holm p \\
\midrule
FEQ vs CCQ & 14 & -0.1392 [-0.5546, 0.3119] & 1.0000 & -0.1253 [-0.6364, 0.4305] & 1.0000 \\
FEQ vs HNQ & 14 & 0.9085 [0.7755, 0.9750] & 0.0003 & 0.8901 [0.6438, 0.9690] & 0.0006 \\
FEQ vs PCQ & 14 & 0.9480 [0.8668, 0.9866] & 0.0003 & 0.8418 [0.4375, 0.9864] & 0.0042 \\
HNQ vs CCQ & 14 & 0.0241 [-0.5919, 0.3897] & 1.0000 & -0.0769 [-0.5491, 0.5416] & 1.0000 \\
PCQ vs CCQ & 14 & -0.2106 [-0.5790, 0.3588] & 1.0000 & 0.0154 [-0.4866, 0.5236] & 1.0000 \\
PCQ vs HNQ & 14 & 0.8160 [0.5836, 0.9471] & 0.0010 & 0.7582 [0.3272, 0.9418] & 0.0106 \\
\bottomrule
\end{tabular}%
}
\end{table}

\begin{figure}[!htbp]\centering
\includegraphics[width=0.82\textwidth]{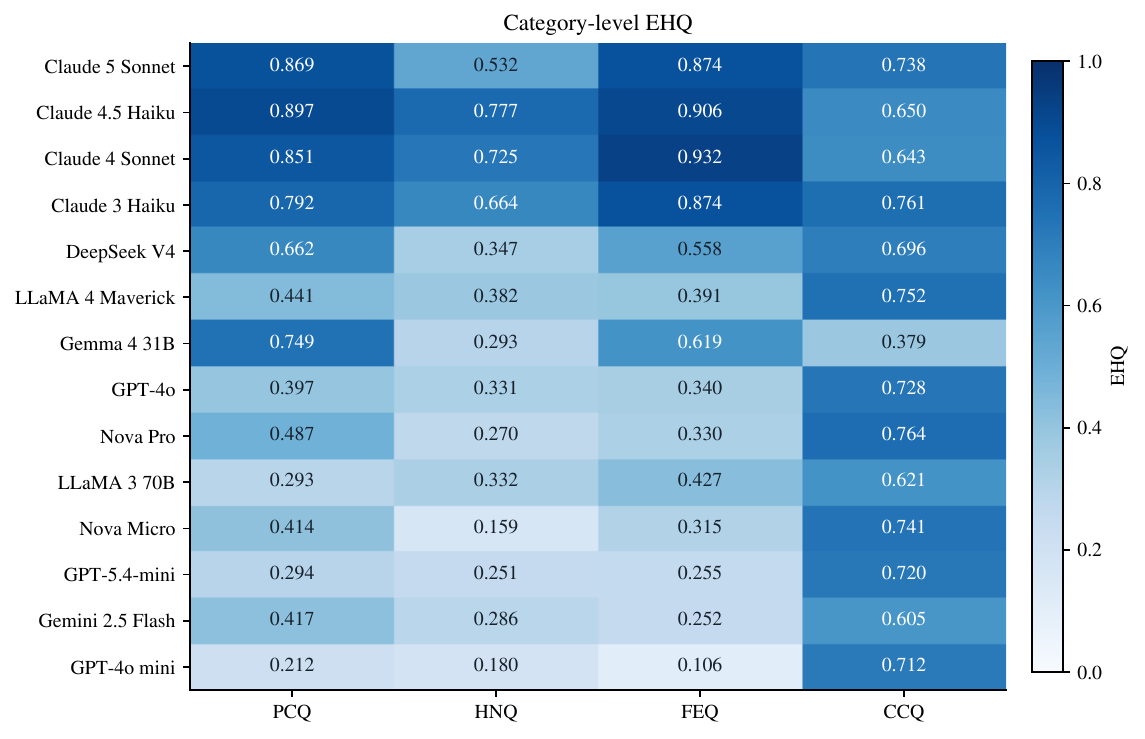}
\caption{Composite EHQ within each question category. FEQ probes fabricated
entities, PCQ post-cutoff events, HNQ source-backed low-accessibility facts, and CCQ questions whose
requested values were withheld from a supplied document. The full per-component breakdown is
Supplementary Table~S14.}
\label{fig:ehq-category-heatmap}
\end{figure}

\paragraph{Response distribution.}
To explain the category and component profiles more directly,
Figure~\ref{fig:ehq-response-distribution} reports how the four response types
divide over scored records; exact counts are Supplementary Table~S6.

Unqualified wrong answers dominate the lower half of the panel, whereas the
four Claude models produce far fewer of them. The split between ABSTAIN and
HEDGE also varies sharply even among models with similar
EHQ\textsubscript{1}. However, human validation shows that the deterministic
classifier identifies HEDGE poorly. We therefore retain the four-class counts
for auditability but do not interpret the within-restraint split as evidence
of distinct behavioral strategies.

Correct confident answers are rare, as the design intends, but not absent. FEQ
and CCQ contribute none by construction because neither category exposes a
gold answer. Instead, $83.1\%$ occur in HNQ, whose source-backed facts were
selected for low expected accessibility rather than guaranteed internal
ignorance. Consequently, some HNQ items remain answerable to current models, a
scope condition revisited in Section~\ref{sec:limitations}. Exact model and
category counts are reported in Supplementary Table~S6.

\begin{figure}[!htbp]\centering
\includegraphics[width=0.86\textwidth]{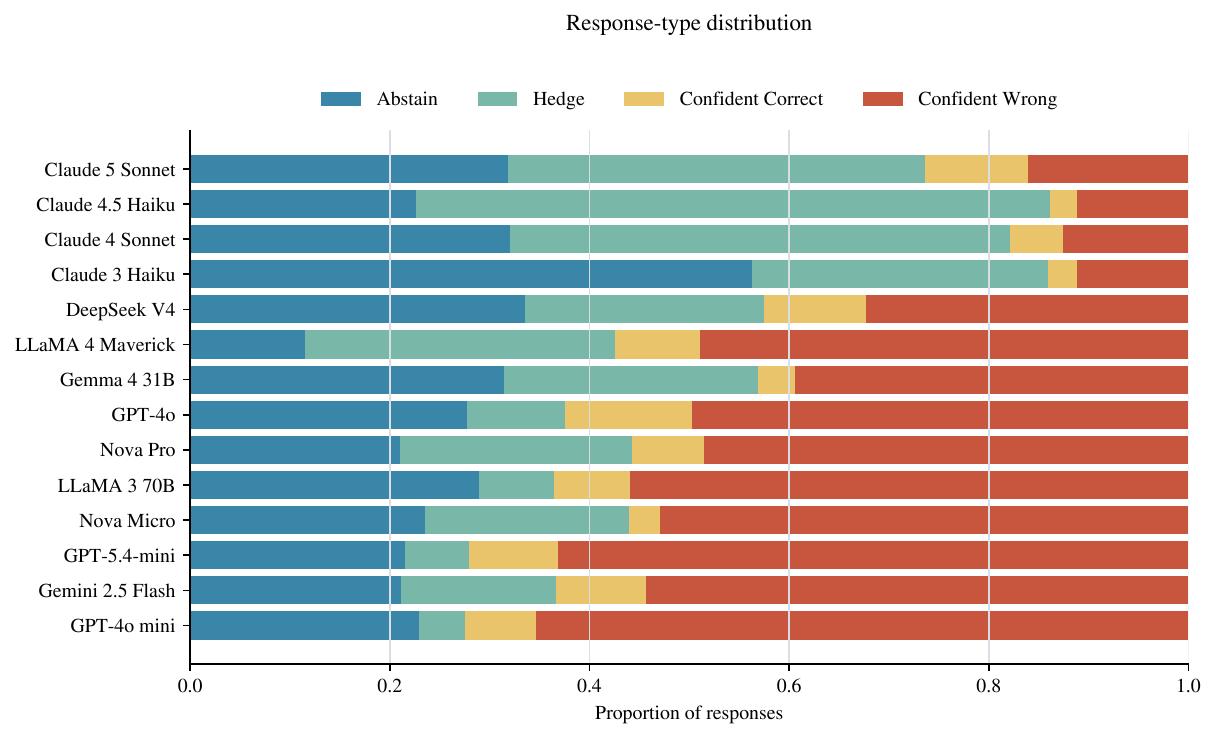}
\caption{How each model's scored responses divide between the four
response types. ABSTAIN and HEDGE are restraint; CONFIDENT WRONG is the legacy
protocol label for an unqualified false assertion. Models are in EHQ order, so the growth of this band
down the panel is the composite's main driver.}
\label{fig:ehq-response-distribution}
\end{figure}

\paragraph{Human validation of the response classifier.}
Table~\ref{tab:classifier-validation-headlines} separates five headline
results from the full class- and model-level diagnostics in Supplementary
Tables~S16--S29. First, independent coders reproduced the codebook closely.
Whereas human--human agreement was high at both resolutions, agreement between
the deterministic classifier and the resolved human reference was only modest
for four classes and stronger for the restraint/substantive distinction. The
rule system therefore supports coarse classification more clearly than it
supports replacing four-class human judgement, particularly for HEDGE.

Consistent with this discrepancy, automation understated both restraint
components in the stratified sample and expanded their spread. Model-level
automated and human-reference scores nevertheless remained strongly
associated. Thus, the validation supports limited comparative stability, not
unbiased absolute levels or a human-corrected full-run ranking.

Moreover, the error was not uniform across model families. A post-hoc
diagnostic found an EHQ1 family-minus-other bias difference of $+0.116$
(exact permutation $p=.021$). Because the contrast is observational, was
specified after inspection, and includes only fourteen models, it indicates
family-associated classifier non-invariance rather than a causal Claude
effect. Partial correlations preserve the family association after controlling
for human-reference score (partial $r=.558$--$.658$), whereas the score
association reverses after controlling for family (partial $r=-.364$ to
$-.217$; Supplementary Table~S25). Thus, score level alone does not account for
the observed family contrast.

Finally, classification also determines which responses enter
EHQ\textsubscript{3}. Classifier error nevertheless returns $23.0\%$ of
human-reference restraint records to the automated substantive pool, so the
definition removes intentional restraint but not classification-boundary error
(Supplementary Tables~S27--S28). Recomputing the composite with the human labels preserved
the same four Claude routes in the first four positions and produced
$\rho=.859$, with a maximum rank shift of three. However, some human-reference
calibration pools are very small; this remains a sparse-sample sensitivity, not
a corrected leaderboard.

\begin{table}[htbp]
\centering
\caption{Headline results from the stratified human validation of the response classifier. Bias is the automated score minus the resolved human-reference score. The family contrast is post hoc and observational. Complete agreement, class-level, model-level, and sensitivity results are reported in Supplementary Tables~S16--S29.}
\label{tab:classifier-validation-headlines}
\small
\renewcommand{\arraystretch}{1.08}
\begin{tabularx}{\linewidth}{>{\raggedright\arraybackslash}p{0.29\linewidth}>{\raggedright\arraybackslash}p{0.25\linewidth}>{\raggedright\arraybackslash}X}
\toprule
Diagnostic & Headline estimate & Interpretation \\
\midrule
Human codebook reproducibility & $\kappa=.913$ (four class); $.940$ (binary) & High reproducibility between independent coders \\
Classifier--human agreement & $\kappa=.414$ (four class); $.667$ (binary) & Stronger support for restraint/substantive classification \\
Mean component bias & EHQ1 $-0.139$; EHQ2 $-0.148$ & Automated labels understate restraint in the validation sample \\
Family diagnostic (EHQ1) & $+0.116$, exact $p=.021$ & Family-associated non-invariance; no causal attribution \\
Human-label composite sensitivity & Spearman $\rho=.859$; maximum rank shift $3$ & Comparative ordering is partly preserved, not human-corrected \\
\bottomrule
\end{tabularx}
\end{table}

\paragraph{Where the calibration evidence sits.}
Because EHQ\textsubscript{3} depends on which responses enter the calibration
set, Figure~\ref{fig:ehq-calibration-sources} reports the decomposition motivating
Section~\ref{subsec:ehq3-validity}: per-model compliance with the report-zero
instruction on abstentions, and the calibration error of the restraint records
EHQ\textsubscript{3} excludes beside the substantive records it retains.

Two facts in that figure carry the argument. First, restraint records contain
essentially no correct answers, so their calibration error is approximately the
mean confidence stated while declining to answer. It therefore measures how
loudly a model expresses confidence during restraint, not how well confidence
tracks correctness. Second, this quantity varies far more across models than
the substantive-answer calibration error with which it was previously pooled.

Restricting EHQ\textsubscript{3} to substantive answers therefore changes scores
materially and in both directions. Against the superseded pooled definition,
EHQ\textsubscript{3} falls by as much as $0.225$ (Nova-Micro) and rises by as
much as $0.145$ (Claude-5-Sonnet). The direction is not arbitrary: a model that
complied with the report-zero instruction contributed near-zero-confidence
restraint records that flattered its pooled ECE, and loses that flattery under
the current definition, while a model that reported high confidence on
abstentions was penalised for it and is no longer.
The per-model restraint accuracies and both score definitions are reported in
Supplementary Table~S24.

\begin{figure}[!htbp]\centering
\includegraphics[width=0.90\textwidth]{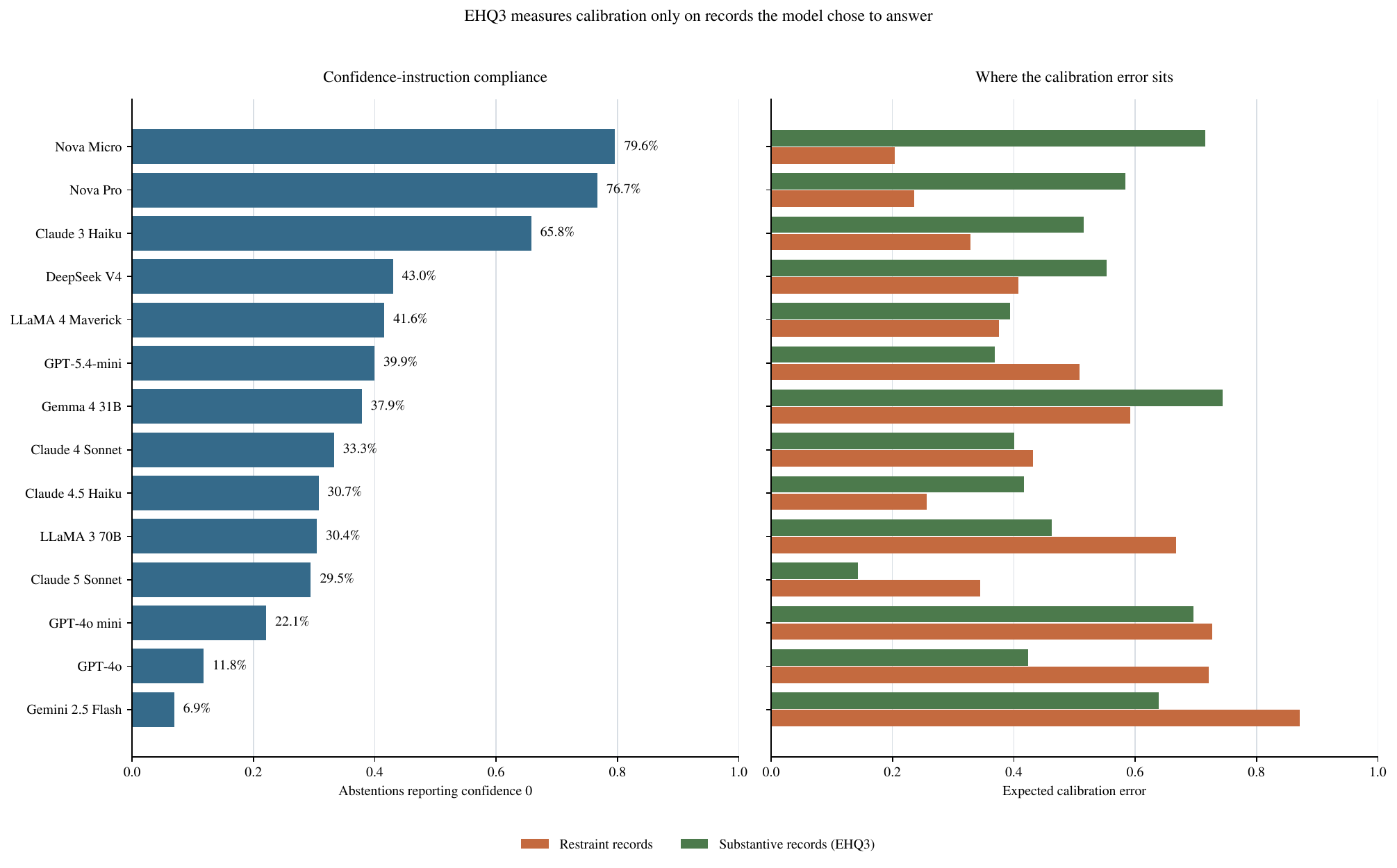}
\caption{Left: the share of each model's abstentions that complied with the
instruction to report confidence 0, an instruction whose correct answer is
known. Right: expected calibration error computed over the restraint records
EHQ\textsubscript{3} now excludes, beside the substantive records it is computed
on. Models are ordered by compliance.}
\label{fig:ehq-calibration-sources}
\end{figure}

As a robustness check, category-balanced scoring confirms that pooled EHQ3 is
partly composition-dependent. Equal weighting changes the composite by at most
$0.060$ and shifts a model by at most three places; notably,
Claude-5-Sonnet moves from first to fourth. Conversely, equal weighting
amplifies noise in categories with very small substantive-answer sets. The
analysis therefore establishes category-composition sensitivity, not the
superiority of either weighting rule, and the official score remains unchanged
(Supplementary Tables~S21--S22).

\subsection{RQ1: does a document-grounded capability floor explain EHQ variation?}
\label{subsec:rq1}

Having established the response profiles, RQ1 addresses a narrower alternative
explanation. Low scores may reflect a
more basic inability to report information that is explicitly available. RQ1
tests that narrow confound, on the same panel and under the same protocol,
by asking each model 259 questions whose answers are present in the document
supplied with them. These items are the released CCQ documents with the
withheld span restored, so the information is available regardless of any
model's knowledge cutoff, and a refusal is scored as incorrect because
declining to report an available value is a failure to deliver it.

Because the probe restores the withheld span in the same source documents, it
also supplies a matched selectivity anchor for CCQ. Across 3,624 gradeable
model--item pairs, models restrain on the missing-span version in $96.6\%$,
answer the restored-span version correctly in $99.2\%$, and pass both sides of
the pair in $95.8\%$. This contrast shows that high CCQ restraint is not simply
blanket refusal when the missing value is supplied.
However, missing-span restraint still uses the deterministic classifier, and
no analogous matched answerable control was run for FEQ, PCQ, or HNQ
(Supplementary Table~S23).

However, the probe produces a floor rather than a spectrum. Pooled accuracy is
$99.2\%$, with only zero to five errors per model. Under the prespecified
shared-rate resampling test, the observed dispersion is not distinguishable
from sampling variation ($p=.30$). This is a failure to detect heterogeneity at
the available sample size, not an equivalence result.

We therefore withhold the capability--EHQ correlation from inferential
reporting: because per-model counts are consistent with a shared rate, the
prespecified analysis gate is triggered. This gate is not cosmetic. Any
correlation would be driven by the same zero-to-five-error variation that the
homogeneity test characterised as sampling noise.

What the probe does establish is the thing RQ1 was posed to settle. Composite
EHQ ranges from $0.31$ to $0.81$ across these fourteen models, a spread $26$
times the capability range. The panel supplies an unusually clean illustration.
GPT-5.4-mini answered all 259 probe items correctly, a result matched only by
Claude-3-Haiku, while placing twelfth of fourteen on epistemic honesty
with an EHQ of $0.41$. Claude-5-Sonnet, first at $0.81$, made one error on the
probe. Whatever separates them, it is not the ability to state a fact that is in
front of them: on that probe no heterogeneity was detected, and the model with the
perfect score is near the bottom of the honesty ordering. The observed EHQ
spread therefore cannot be explained by capability differences detectable with
this deliberately easy probe.

The claim is bounded by what the probe measures, and we state the bound rather
than let it be inferred. Answering from a supplied document is deliberately
easy; it was chosen to be a floor. A capability measure that did separate the
panel, such as reasoning depth, long-horizon accuracy, or performance on a demanding
benchmark, might well correlate with EHQ. However, this study does not test that possibility.
We considered substituting published MMLU-Pro or GPQA results, but did not do
so: exact-route coverage is incomplete, and available values often come from
reasoning, tool-use, or other inference conditions that differ from the
non-reasoning protocol enforced here. Combining those values post hoc would
introduce both non-random missingness and protocol heterogeneity. A future
confirmatory test should administer one discriminating capability battery
through the same endpoints and inference settings used for EHQ.
What is ruled out is the narrower and more immediate confound: that low EHQ
scores merely reflect an inability to extract explicit values from the short
documents used in this probe.

\subsection{RQ2: do newer members of prespecified model families differ in EHQ from their older counterparts?}
\label{subsec:rq2-results}
RQ2 shifts from the cross-sectional panel to within-family change. Four
registered old/new pairs had both members clear the eligibility gates.
Figure~\ref{fig:ehq-pair-differences} shows the differences with paired
bootstrap intervals; numerical summaries are Supplementary Tables~S9--S10.

All four newer models score higher, every paired interval excludes zero, and
the mean difference is $+0.052$. The largest gain is GPT-5.4-mini over
GPT-4o-mini ($+0.093$); the remaining pair-level estimates and intervals are
shown in Figure~\ref{fig:ehq-pair-differences} and Supplementary
Tables~S9--S10.

However, the exact sign-permutation test returns $p=.125$. With only four
pairs, this is also the smallest attainable two-sided $p$-value: four
improvements and no declines are the strongest evidence the design admits, yet
they cannot reach conventional significance. We therefore report a consistent
direction without describing a population-level generational effect as
statistically significant.

Four of the eight registered pairs contributed nothing, each for a documented
reason: DeepSeek-V3 is no longer servable on its provider, the two Gemini pairs
have new members whose reasoning cannot be disabled under this protocol, and
GPT-5.5 does not return responses reliably enough to score. These are recorded
in the model registry with their evidence.

\begin{figure}[htbp]\centering
\includegraphics[width=\textwidth]{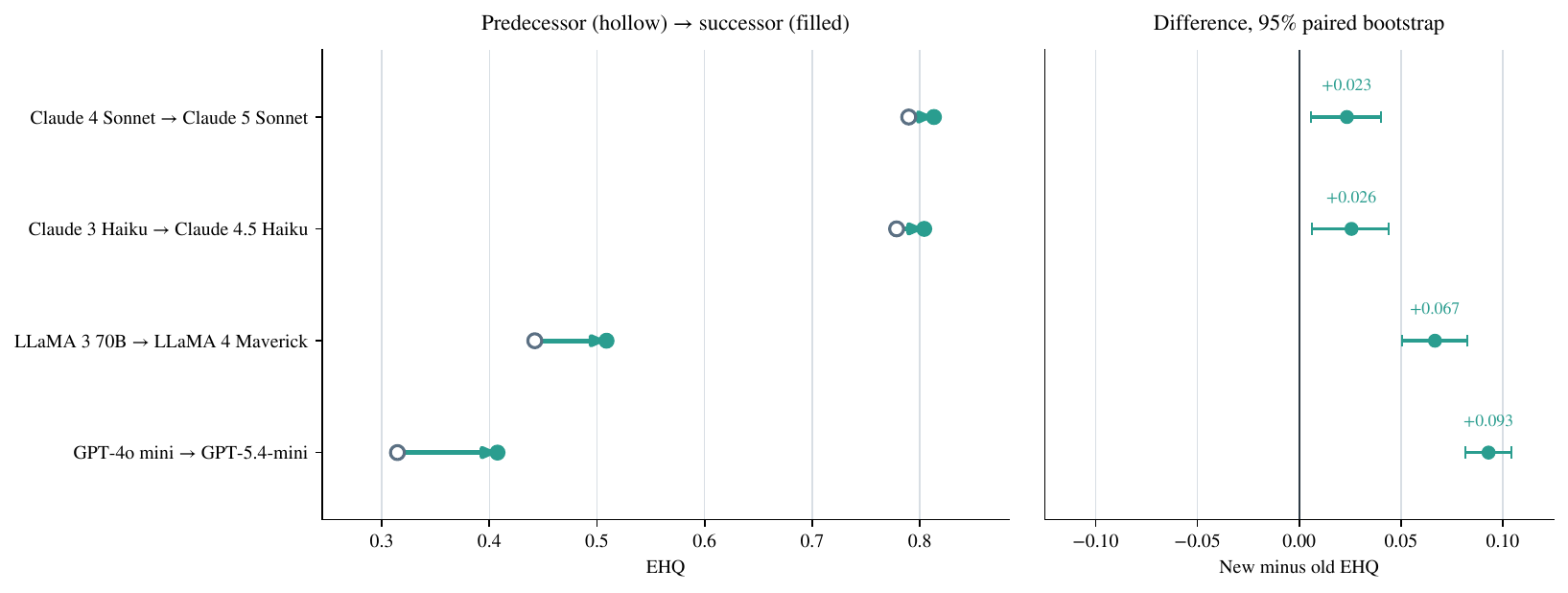}
\caption{Left: each registered generational pair on the EHQ scale, from
predecessor (hollow) to successor (filled). Right: the same comparison as a
new-minus-old difference with its 95\% paired bootstrap
interval. Pairs are ordered by the size of the gain.}
\label{fig:ehq-pair-differences}
\end{figure}

\subsection{RQ3: how are EHQ1, EHQ2, and EHQ3 related across models?}
\label{subsec:rq3-results}

Finally, RQ3 asks how the sub-scores move together. Table~\ref{tab:ehq-correlations} reports Pearson and Spearman
coefficients between the three components over the fourteen panel models, with
bootstrap intervals, permutation $p$-values and Holm-adjusted $p$-values.

EHQ\textsubscript{3} shows no statistically detectable association with either
other component in this fourteen-model panel. The Pearson estimates are
$r=.333$ for EHQ1 and $r=.420$ for EHQ2, with both bootstrap intervals
including zero (Table~\ref{tab:ehq-correlations}). Spearman estimates in that
table and the Fisher-$z$ sensitivity in the supplement lead to the same
conclusion. However, the intervals remain compatible
with moderate positive effects. The result therefore does not establish
independence; it shows only that restraint did not reliably predict
substantive-answer calibration in this small panel.

\paragraph{Dataset-dependent overlap between EHQ1 and EHQ2.}
EHQ\textsubscript{1} and EHQ\textsubscript{2} correlate at $r = +0.990$
$[0.977, 0.997]$ in this panel. Their shared behavioral base follows from the
definitions, but the magnitude of the correlation is empirical and
dataset-dependent. Writing $n$ for the scored records and $CC$, $CW$ for
\texttt{CONFIDENT\_CORRECT} and \texttt{CONFIDENT\_WRONG} counts,
\begin{equation*}
\mathrm{EHQ}_1 = 1 - \frac{CC + CW}{n},
\qquad
\mathrm{EHQ}_2 = 1 - \frac{CW}{n},
\qquad\text{so}\qquad
\mathrm{EHQ}_2 - \mathrm{EHQ}_1 = \frac{CC}{n}.
\end{equation*}
The identity establishes algebraic nesting, not behavioral equivalence.
EHQ\textsubscript{1} asks whether a model refrains from a definitive substantive
answer, whereas EHQ\textsubscript{2} asks whether it avoids a definitive
falsehood; they differ exactly on \texttt{CONFIDENT\_CORRECT} responses. In EHQ-3000 that
rate is small, varies little across models, and is structurally zero on FEQ and
CCQ. On half the dataset the two scores are therefore identical. A dataset offering more
frequent and more heterogeneous opportunities for correct substantive answers
could separate them substantially. The present result supports neither
behavioral equivalence nor independent latent dimensions; it shows strong
collinearity under this dataset's composition.

At the category level, FEQ and CCQ produce exact identity and PCQ offers little
separation. HNQ is the only category with a material gap between the two
criteria and a meaningfully different rank ordering. Thus, the criteria can
distinguish model behaviour when correct substantive answers occur often
enough, although this does not establish independent latent dimensions
(Supplementary Table~S15).

Accordingly, the consequence for the composite is worth stating plainly. Substituting the
identity above into the frozen $0.30/0.45/0.25$ weighting gives
$EHQ=0.75EHQ_1+0.45(CC/n)+0.25EHQ_3$. The composite therefore rewards
restraint, correct unqualified answers, and calibration; describing it as only
$0.75$ restraint and $0.25$ calibration would omit the confident-correct term.
As a robustness check, dropping either nested control criterion moves no model
by more than two ranks, while equal weighting of restraint and calibration
moves models by at most four. Full estimates are reported in Supplementary
Table~S11.

Taken together, the sensitivity results show that the present model ordering is
robust to the strong dataset-specific overlap: no conclusion in this paper
depends on which nested control criterion is retained. EHQ\textsubscript{1}
reports response restraint, EHQ\textsubscript{2} reports avoidance of unqualified
false assertions, and EHQ\textsubscript{3} reports confidence--accuracy alignment on
substantive answers. The observed correlations suggest that calibration contributes
information not captured by restraint, but the panel is too small to establish
statistical independence.

\begin{table}[htbp]
\centering
\caption{Correlations between the three EHQ components across the panel. The unit of inference is the model, not the item. Intervals are bootstrap percentile intervals over models and p-values come from permutation tests, Holm-adjusted within the Pearson and Spearman families of three tests each.}
\label{tab:ehq-correlations}
\small
\resizebox{\ifdim\width>\linewidth\linewidth\else\width\fi}{!}{%
\begin{tabular}{lrrrrrrr}
\toprule
Comparison & n & Pearson r [95\% CI] & p & Holm p & Spearman rho [95\% CI] & p & Holm p \\
\midrule
EHQ1 vs EHQ2 & 14 & 0.9904 [0.9769, 0.9972] & < 0.0001 & 0.0001 & 0.9868 [0.8914, 1.0000] & < 0.0001 & 0.0001 \\
EHQ1 vs EHQ3 & 14 & 0.3329 [-0.1990, 0.6685] & 0.2454 & 0.2649 & 0.1692 [-0.4286, 0.6629] & 0.5629 & 0.8028 \\
EHQ2 vs EHQ3 & 14 & 0.4199 [-0.1153, 0.7397] & 0.1324 & 0.2649 & 0.2440 [-0.3529, 0.7073] & 0.4014 & 0.8028 \\
\bottomrule
\end{tabular}%
}
\end{table}

\fi

\section{Discussion} \label{sec:discussion}

Across identical items and inference settings, EHQ separates the analysed
models by a factor of $2.6$ and reveals trade-offs that an accuracy score would
not show. The four Claude routes occupy the first four positions under the
frozen automated composite. Importantly, the human-labelled subsample preserves
those four routes in its first four composite positions, although in a
different order ($r=.923$, $\rho=.859$; maximum shift, three ranks). The main
finding is therefore not a leaderboard but the structure beneath it: restraint,
calibration, and boundary type come apart in distinct ways. Classifier error
limits absolute component estimates and some family comparisons, but it does
not erase that comparative pattern in the validation sample.

\subsection{Behavioral implications across components, categories, and generations}

\paragraph{Restraint and calibration are separate, and models trade them.}

Although EHQ\textsubscript{3} shows no statistically detectable association with restraint
(Section~\ref{subsec:rq3-results}), the panel contains clear instances of
each combination. Claude-4.5-Haiku declines to answer more often than any other
model ($0.8613$) and is unremarkable at calibration ($0.5831$). GPT-5.4-mini
does the opposite: it answers nearly everything ($\mathrm{EHQ}_1 = 0.2797$,
second lowest) while ranking second on calibration ($0.6309$). Claude-5-Sonnet
takes first place not by refusing most often (it refuses least among the four
Claude models), but by being much better calibrated on what it does assert
($0.8574$).

These are recognisably different deployment profiles. A model that abstains
freely is safe in the sense that it emits few unqualified false assertions, but it is
also less useful, and its restraint says nothing about whether the confidence
attached to its remaining answers can be trusted. A model that answers
everything with well-tracked confidence is the opposite bargain: more
assertions, but a usable signal for deciding which to check. Neither profile is
strictly better, and a single composite necessarily obscures the choice. This
is the practical case for reporting the components alongside EHQ rather than in
place of it.

\paragraph{Knowledge- and context-boundary profiles may differ.}

Whereas the preceding comparison concerns components, the stronger observed
dissociation is between categories
(Section~\ref{subsec:category-profiles}). Performance on FEQ, PCQ and HNQ moves
together ($r = +0.82$ to $+0.95$). CCQ point estimates are smaller
($r = -0.21$ to $+0.02$), but their intervals remain wide. Nova-Pro is tenth of
fourteen on FEQ and first on CCQ; Claude-4-Sonnet is first on FEQ and eleventh
on CCQ.

The distinction the data draws is between two objects of judgement. FEQ, PCQ
and HNQ ask a model to recognise the limits of what it has learned: whether an
entity exists, whether an event postdates its training, whether a fact is too
obscure to have been acquired. CCQ asks it to recognise the limits of a
document placed in front of it. The first is a judgement about the model's own
knowledge; the second is a judgement about a text. The observed pattern is
consistent with partially distinct competences, but does not prove independence.

That has a direct consequence for retrieval-augmented deployment, which is
where most production systems now sit. Choosing a model because it declines to
invent nonexistent citations does not predict that it will decline to invent a
value the retrieved document does not contain. The weak observed association is
not precise enough to substitute one test for the other, so a system that needs
the second ability should be evaluated on it directly.

\paragraph{All observed newer models score higher, but inference remains inconclusive.}

Across the eligible generational pairs, all four newer models improved, with per-pair intervals
excluding zero and a mean gain of $+0.052$
(Section~\ref{subsec:rq2-results}). The largest gain, GPT-5.4-mini over
GPT-4o-mini at $+0.093$, still leaves the newer model twelfth of fourteen: the
family improved without moving out of the lower half of the panel. Epistemic
honesty in these four pairs is consistent with generational improvement, but
the design is too small to establish a population-level trend.

The evidence is also weaker than four-for-four suggests. With four pairs the
exact sign test cannot return a $p$ below $0.125$, so the strongest possible
outcome is still conventionally non-significant. Four of the eight registered
pairs contributed nothing, each for a reason recorded in the model registry,
and three of those four failed for provider or protocol reasons rather than
model behaviour. A larger panel would not merely tighten this estimate; it is
the only thing that would make it testable.

\subsection{Interpreting EHQ and its measurement}

\paragraph{What the composite does and does not measure.}

Two results constrain how EHQ should be read. First, EHQ\textsubscript{1} and
EHQ\textsubscript{2} target different control questions, namely whether the model
withholds a definitive answer and whether it avoids a definitive falsehood, but
are algebraically nested and are identical on the half of EHQ-3000 where a
\texttt{CONFIDENT\_CORRECT} outcome is structurally unavailable. Their $r=+0.990$
association is therefore specific to the present dataset and panel, not proof
that the intended behaviors are equivalent. The exact composite decomposition
is $0.75EHQ_1+0.45(CC/n)+0.25EHQ_3$, and dropping either nested control
sub-score moves no model by more than two ranks
(Section~\ref{subsec:rq3-results}). Only HNQ supplies material within-category
separation in this realisation (Supplementary Table~S15). The three reported sub-scores retain
distinct interpretations without being claimed as three independent latent
dimensions.

The human-reference sensitivity analysis provides a direct, though limited,
check on the composite. Replacing automated labels within the stratified sample
preserves strong score and rank association, retains the same four routes in
the first four positions, and moves no route by more than three ranks. This
supports bounded comparative use of EHQ in the present panel. It does not
validate the automated absolute levels, prove family-rank invariance, or replace
a full human-labelled rerun because each model contributes only 40 sampled
responses and some human-reference EHQ3 pools are very small.

Second, the composite is not a restatement of the document-grounded floor skill
measured here. Every model in the panel answers those questions at $99.2\%$, with no detectable
between-model heterogeneity (Section~\ref{subsec:rq1}), while EHQ varies over a range $26$ times
wider. The sharpest case is GPT-5.4-mini, which answers all $259$ probe items
correctly and places twelfth of fourteen on epistemic honesty. The observed EHQ
variation is therefore not explained by performance on this near-ceiling
probe. The bound on that claim matters and we state it: the probe is a
floor, chosen to be easy, and a capability measure that did separate the panel
might well correlate with EHQ.

The matched CCQ contrast nevertheless addresses one narrower construct concern.
Models usually withhold the missing value and answer when the same value is
restored, which is evidence of selective context-boundary behavior rather than
uniform refusal. It does not solve the general over-refusal problem. On a
boundary-only benchmark, blanket restraint maximises EHQ1 and EHQ2, although it
leaves substantive-only EHQ3 and therefore the composite undefined when no
substantive answer is produced. Consequently, EHQ must be paired with utility
or answerable-set performance in deployment. The present study supplies that
paired control only for CCQ; matched real-entity, pre-cutoff, and common-fact
controls remain necessary for FEQ, PCQ, and HNQ.

\paragraph{Provider and elicitation failures alter the measurement.}

Three of this study's findings are about the measurement rather than the
models, and we report them as results because a reader repeating the work will
meet them.

A third of one model's answers were truncated mid-sentence by the provider
while returning success, which the classifier reads as substantive content;
scored as published, that model ranks last, and under the reading its surviving
fragments support it ranks sixth. Its position is undetermined, and no
inspection of its scores would have revealed that.

The confidence value elicited on a restraint response is not a comparable
quantity across models. Compliance with an explicit, verifiable instruction
ranges from $6.9\%$ to $79.6\%$ on all scored abstentions, and among
non-compliant models eleven treat
the scale as a switch while three use it as a scale
(Section~\ref{subsec:ehq3-validity}). Pooling calibration error over such
records combines numbers that do not denote the same thing, which is why
EHQ\textsubscript{3} is now computed over substantive answers only.

Finally, the non-reasoning condition this protocol requires cannot be requested from
the gateway used here; it can only be checked afterwards from provider-reported
token accounting, and two registered routes fail that check by a wide margin.
Evaluations that assume a configuration flag has taken effect, without
verifying it per response, may be measuring a different protocol than the one
they describe.

\subsection{Limitations} \label{sec:limitations}

\noindent\textbf{Scope and benchmark boundaries.}
EHQ is a behavioral measure on English, single-turn prompts with verbalized
confidence; it does not observe internal knowledge states or identify causal
mechanisms. PCQ depends on claim-level novelty and documented model cutoffs,
whereas HNQ denotes low expected accessibility rather than guaranteed
ignorance. The secondary human review covers 20\% of PCQ/HNQ and 10\% of
EHQ-3000 overall, reducing but not eliminating source or annotation error in
those strata; FEQ and CCQ did not receive that second pass. DeepSeek-V4 lacked
an exact public cutoff, so its PCQ eligibility uses a named-human conservative
upper-bound adjudication rather than an asserted training date.

\noindent\textbf{Controls and temporal comparability.}
The restored-span probe is a matched answerable control for CCQ, but FEQ, PCQ,
and HNQ lack corresponding real-entity, pre-cutoff, and common-fact controls.
Consequently, restraint in those categories may reflect appropriate boundary
recognition, general over-refusal, or both. PCQ also guarantees eligibility by
placing claims after the latest registered boundary, but it does not equalise
cutoff-to-event distance across models; RQ2 therefore combines generation change
with temporal-distance differences. PCQ should be maintained as a versioned
release, and future versions should add preregistered matched controls and
threshold items on which correct definitive answers are possible but
non-trivial. These additions would test utility and separate EHQ1 from EHQ2 more
effectively without changing the frozen estimand post hoc.

\noindent\textbf{Classification and confidence measurement.}
The deterministic classifier is reproducible but is not a validated substitute
for human judgement. Each model contributes only 40 validation records, the
family diagnostic is observational and post hoc, and the 25 adjudicated
disagreements used disclosed AI decision support. The validation neither
corrects the 3,000-item scores nor establishes a causal family effect; any
revised classifier requires a new, untouched human-coded sample. Likewise, an
elicited confidence number may be a differently scaled self-report rather than
a comparable probability across models. Restricting EHQ3 to substantive answers
removes the clearest mismatch, but robustness to confidence-prompt wording must
be re-established under this protocol~\cite{webson2022prompt}.

\noindent\textbf{Provider and protocol effects.}
Technical failures are excluded from all components, whereas ambiguous but
technically successful confidence responses remain in EHQ1 and EHQ2 and are
excluded from EHQ3; neither is imputed. Provider-side truncation is more
consequential: sensitivity analysis leaves thirteen routes essentially stable
but makes Gemini-2.5-Pro's standing undetermined. It also leaves a lower-panel
uncertainty region: Nova-Micro ranks 11--13, GPT-5.4-mini 12--14, and
Gemini-2.5-Flash 12--13 (Supplementary Table~S13). The non-reasoning condition can
only be checked from provider-reported token accounting, so routes without that
accounting remain unverified. Finally, EHQ3 values under the current
substantive-only protocol are not comparable with values from the superseded
pooled definition; the protocol definition is part of the estimand.

\noindent\textbf{Composition and statistical power.}
Because each model determines which responses enter EHQ3, pooled calibration
induces model-specific category weights. Equal category weighting remains
strongly associated with the official score but moves one model by three ranks
and can amplify small cells, so EHQ3 should be reported with per-category pool
sizes and any deployment-specific weights should be justified in advance.
HNQ supplies $83.1\%$ of all \texttt{CONFIDENT\_CORRECT} records and therefore
most observed separation between EHQ1 and EHQ2; however, removing HNQ leaves
official EHQ highly correlated ($r=.987$) and changes no model by more than two
ranks (Supplementary Table~S12). Model-level inference is also limited by the
near-ceiling RQ1 probe and only four complete generational pairs. Accordingly,
the present results characterize this frozen panel and protocol rather than a
population of model families or deployment settings.

\section{Conclusion} \label{sec:conclusion}

We introduced the Epistemic Honesty Quotient, a measure of how a model behaves
on questions whose answers are structurally unavailable, post-cutoff, withheld
from context, or deliberately low-accessibility, together with EHQ-3000, a
balanced benchmark of three thousand such questions across four categories, and a framework that
verifies endpoint identity, enforces its inference conditions per response, and
seals every run against a hash catalogue. Fifteen routes completed the
confirmatory evaluation; fourteen form the analysed panel.

EHQ varies far more across the panel than performance on the near-ceiling
document probe, and its structure matters: restraint and calibration show no
statistically detectable association, while knowledge- and context-boundary
profiles are only weakly associated. All four observed newer models improve by
a mean of $0.05$, although four pairs cannot establish a general trend. The
matched CCQ contrast further shows category-specific selectivity because models
usually answer when the withheld value is restored; equivalent controls remain
necessary for the other categories.

The audit also defines the evidential boundary. Provider truncation can make a
standing undetermined, confidence after restraint is not comparable across
models, and a non-reasoning condition must be verified per response. Human
coders reproduced the four-class codebook, but the deterministic classifier was
more reliable for restraint versus substantive responding than for individual
labels. Human-label recomputation preserved the broad composite ordering while
showing family-associated component error. EHQ therefore supports bounded
comparative interpretation more strongly than unqualified claims about
absolute levels or family positions.

Finally, the dataset, model registry, frozen run artifacts, and analysis code form one
hash-bound release package from which every figure can be regenerated.

\section*{Data and Code Availability}

The EHQ-3000 dataset and the provider-neutral EHQ evaluation framework are publicly available at \url{https://github.com/senolali/EHQ}. The exact dataset release used in this study can be downloaded directly from \url{https://raw.githubusercontent.com/senolali/EHQ/main/data/releases/EHQ-3000.json}. The framework is also distributed as the \texttt{ehq} Python package through PyPI at \url{https://pypi.org/project/ehq/}.

The repository provides versioned source code, configuration templates, analysis scripts, release metadata, and instructions for reproducing the evaluation and generating the associated analyses, tables, and figures. Machine-readable implementation identifiers are documented in the release metadata rather than the narrative manuscript. Model-provider credentials, proprietary endpoints, and third-party services are not redistributed.


\bibliographystyle{plain} 

\begin{thebibliography}{99}

\bibitem{wei2022chain} Jason Wei et al. \newblock Chain-of-thought prompting elicits reasoning in large language models. \newblock \emph{Advances in Neural Information Processing Systems}, 35:24824--24837, 2022.

\bibitem{kojima2022large} Takeshi Kojima et al. \newblock Large language models are zero-shot reasoners. \newblock \emph{Advances in Neural Information Processing Systems}, 35:22199--22213, 2022.

\bibitem{brown2020language} Tom Brown et al. \newblock Language models are few-shot learners. \newblock \emph{Advances in Neural Information Processing Systems}, 33:1877--1901, 2020.

\bibitem{turpin2023language} Miles Turpin, Julian Michael, Ethan Perez, and Samuel Bowman. \newblock Language models don't always say what they think: Unfaithful   explanations in chain-of-thought prompting. \newblock \emph{Advances in Neural Information Processing Systems}, 36:74952--74965, 2023.

\bibitem{maynez2020faithfulness} Joshua Maynez et al. \newblock On faithfulness and factuality in abstractive summarization. \newblock In \emph{Proceedings of ACL}, pages 1906--1919, 2020.

\bibitem{ji2023survey} Ziwei Ji et al. \newblock Survey of hallucination in natural language generation. \newblock \emph{ACM Computing Surveys}, 55(12):1--38, 2023.

\bibitem{zhang2023siren} Yue Zhang et al. \newblock Siren's song in the AI ocean: A survey on hallucination in large   language models. \newblock \emph{arXiv preprint arXiv:2309.01219}, 2023.

\bibitem{truthfulqa} Stephanie Lin, Jacob Hilton, and Owain Evans. \newblock TruthfulQA: Measuring how models mimic human falsehoods. \newblock In \emph{Proceedings of ACL}, pages 3214--3252, 2022.

\bibitem{es2023ragas} Shahul Es et al. \newblock RAGAS: Automated evaluation of retrieval augmented generation. \newblock In \emph{Proceedings of EACL}, 2024.

\bibitem{min2023factscore} Sewon Min et al. \newblock FActScore: Fine-grained atomic evaluation of factual precision in long-form text generation. \newblock \emph{arXiv preprint arXiv:2305.14251}, 2023.

\bibitem{guo2017calibration} Chuan Guo, Geoff Pleiss, Yu Sun, and Kilian Q.\ Weinberger. \newblock On calibration of modern neural networks. \newblock In \emph{Proceedings of ICML}, pages 1321--1330, 2017.

\bibitem{kadavath2022language} Saurav Kadavath et al. \newblock Language models (mostly) know what they know. \newblock \emph{arXiv preprint arXiv:2207.05221}, 2022.

\bibitem{xiong2024can} Miao Xiong et al. \newblock Can LLMs express their uncertainty? An empirical evaluation of   confidence elicitation in LLMs. \newblock In \emph{Proceedings of ICLR}, 2024.

\bibitem{feng2024don} Shangbin Feng et al. \newblock Don't hallucinate, abstain: Identifying LLM knowledge gaps via   multi-LLM collaboration. \newblock In \emph{Proceedings of ACL}, 2024.

\bibitem{cheng2024can} Qinyuan Cheng et al. \newblock Can AI assistants know what they don't know? \newblock In \emph{Proceedings of the 41st International Conference on Machine Learning (ICML)}, volume 235 of \emph{Proceedings of Machine Learning Research}, pages 8184--8202, 2024.

\bibitem{yang2023alignment} Yuqing Yang, Ethan Chern, Xipeng Qiu, Graham Neubig, and Pengfei Liu. \newblock Alignment for honesty. \newblock \emph{arXiv preprint arXiv:2312.07000}, 2023.

\bibitem{yin2023large} Zhangyue Yin et al. \newblock Do large language models know what they don't know? \newblock In \emph{Findings of ACL}, pages 8653--8665, 2023.

\bibitem{amayuelas2023knowledge} Alfonso Amayuelas et al. \newblock Knowledge of knowledge: Exploring known-unknowns uncertainty with   large language models. \newblock \emph{arXiv preprint arXiv:2305.13712}, 2023.

\bibitem{liang2022holistic} Percy Liang et al. \newblock Holistic evaluation of language models. \newblock \emph{arXiv preprint arXiv:2211.09110}, 2022.

\bibitem{kelly2025epistemic}
Matthew Kelly.
\newblock The Epistemic Suite: A Post-Foundational Diagnostic Methodology
  for Assessing AI Knowledge Claims.
\newblock \emph{arXiv preprint arXiv:2510.24721}, 2025.

\bibitem{tripathi2025confidence}
Sahil Tripathi, Md Tabrez Nafis, Imran Hussain, and Jiechao Gao.
\newblock The Confidence Paradox: Can LLM Know When It's Wrong.
\newblock \emph{arXiv preprint arXiv:2506.23464}, 2025.

\bibitem{srivastava2022beyond} Aarohi Srivastava et al. \newblock Beyond the imitation game: Quantifying and extrapolating the   capabilities of language models. \newblock \emph{arXiv preprint arXiv:2206.04615}, 2022.

\bibitem{senol2026measuring} Ali {\c{S}}enol, Garima Agrawal, and Huan Liu. \newblock Measuring Reasoning Quality in LLMs: A Multi-Dimensional Behavioral Framework. \newblock \emph{arXiv preprint arXiv:2605.24661}, 2026.

\bibitem{nelson1990metamemory} Thomas O.\ Nelson and Louis Narens. \newblock Metamemory: A theoretical framework and new findings. \newblock \emph{Psychology of Learning and Motivation}, 26:125--173, 1990.

\bibitem{flavell1979metacognition} John H.\ Flavell. \newblock Metacognition and cognitive monitoring: A new area of   cognitive-developmental inquiry. \newblock \emph{American Psychologist}, 34(10):906--911, 1979.

\bibitem{lichtenstein1977training} Sarah Lichtenstein and Baruch Fischhoff. \newblock Do those who know more also know more about how much they know? \newblock \emph{Organizational Behavior and Human Performance},   20(2):159--183, 1977.

\bibitem{murphy1977reliability} Allan H.\ Murphy and Robert L.\ Winkler. \newblock Reliability of subjective probability forecasts of precipitation   and temperature. \newblock \emph{Journal of the Royal Statistical Society: Series C},   26(1):41--47, 1977.

\bibitem{kruger1999unskilled} Justin Kruger and David Dunning. \newblock Unskilled and unaware of it: How difficulties in recognizing one's   own incompetence lead to inflated self-assessments. \newblock \emph{Journal of Personality and Social Psychology},   77(6):1121--1134, 1999.

\bibitem{webson2022prompt} Albert Webson and Ellie Pavlick. \newblock Do prompt-based models really understand the meaning of their prompts? \newblock In \emph{Proceedings of NAACL}, pages 2300--2344, 2022.

\bibitem{devilling2025polite} Bentley DeVilling. \newblock The Polite Liar: Epistemic Pathology in Language Models. \newblock \emph{arXiv preprint arXiv:2511.07477}, 2025.


\bibitem{kalai2025why} Adam Tauman Kalai, Ofir Nachum, Santosh~S. Vempala, and Edwin Zhang. \newblock Why Language Models Hallucinate. \newblock \emph{arXiv preprint arXiv:2509.04664}, 2025.

\end{thebibliography}

\end{document}


\maketitle

All numerical tables are bound directly from the same verified publication
package as the main manuscript. This supplement holds detailed audit and
diagnostic material removed from the main text for readability; no analysis or
result is changed.

\section{Prior-work comparison}

\begin{table}[htbp]
\centering
\caption{Comparison of prior work with EHQ. A check mark denotes explicit
measurement, a circle partial coverage, and a cross no coverage. ERR:
Epistemic Restraint Rate; HR: Hallucination Resistance; CA: calibration on the
unknown subset. Independence from correctness is not included as a property:
the present floor probe does not establish it. The general multidimensional
framework in the \c{S}enol et al. row is included as a complementary
predecessor, not because it targets any of the four boundary categories.}
\label{tab:s-gap-analysis}
\scriptsize
\resizebox{\textwidth}{!}{%
\begin{tabular}{lccccccc}
\toprule
Study & ERR & HR & CA & FEQ & PCQ & HNQ & CCQ \\
\midrule
Guo et al.~\cite{sguo2017} & $\times$ & $\times$ & $\circ$ & $\times$ & $\times$ & $\times$ & $\times$ \\
Kadavath et al.~\cite{skadavath2022} & $\circ$ & $\circ$ & $\circ$ & $\times$ & $\times$ & $\times$ & $\times$ \\
Lin et al.~\cite{slin2022} & $\times$ & $\circ$ & $\times$ & $\circ$ & $\times$ & $\times$ & $\times$ \\
Xiong et al.~\cite{sxiong2024} & $\times$ & $\circ$ & $\circ$ & $\times$ & $\times$ & $\times$ & $\times$ \\
Feng et al.~\cite{sfeng2024} & $\circ$ & $\circ$ & $\times$ & $\times$ & $\times$ & $\times$ & $\times$ \\
Yin et al.~\cite{syin2023} & $\circ$ & $\circ$ & $\times$ & $\times$ & $\circ$ & $\times$ & $\times$ \\
Amayuelas et al.~\cite{samayuelas2023} & $\times$ & $\circ$ & $\times$ & $\times$ & $\circ$ & $\times$ & $\times$ \\
Cheng et al.~\cite{scheng2024} & $\circ$ & $\times$ & $\times$ & $\times$ & $\times$ & $\times$ & $\times$ \\
\c{S}enol et al.~\cite{ssenol2026} & $\times$ & $\times$ & $\times$ & $\times$ & $\times$ & $\times$ & $\times$ \\
\midrule
\textbf{This work} & $\checkmark$ & $\checkmark$ & $\checkmark$ & $\checkmark$ & $\checkmark$ & $\checkmark$ & $\checkmark$ \\
\bottomrule
\end{tabular}}
\end{table}

\section{Registry and eligibility}
\ehqtable{model_registry}
\ehqtable{analysis_exclusions}
\clearpage

\section{Scoring coverage}
\ehqtable{coverage_by_model}
\ehqtable{coverage_exclusions}
\clearpage

\section{Response and calibration diagnostics}
\ehqtable{response_distribution}
\ehqtable{calibration_sources}
\ehqauxtable{abstention_confidence_compliance}
\clearpage

\section{Generational-pair analysis}
\ehqtable{rq2_paired_summary}
\ehqtable{pair_differences}
\clearpage

\section{Sensitivity analyses}
\ehqtable{composite_weight_sensitivity}
\ehqtable{construct_scope_sensitivity}

The truncation audit counts a record as suspected when reported completion
tokens reach the configured answer budget or when the text ends on a function
word, dangling separator, or inside an unclosed quotation or bracket. Absence
of sentence-final punctuation alone is recorded as a weaker model-level signal,
not used to flag a record. The panel median for that weaker signal was $0.4\%$,
and twelve of fifteen routes had at most eleven suspected records. The third
route above that threshold, Claude-5-Sonnet, had 18 ($0.6\%$). Much larger
rates occurred for Gemini-2.5-Flash (239/3,000) and Gemini-2.5-Pro (997/3,000, of
which 905 were classified CONFIDENT\_WRONG). No Gemini-2.5-Pro answer reached
the configured 512-token budget and its maximum observed completion was 270
tokens, so the provider-side mechanism cannot be identified from this run.
Table~S13 therefore reports the published score and two explicit bounds rather
than a corrected score.

\ehqauxtable{truncation_sensitivity}
\clearpage

\section{Full category-level scores}
\ehqtable{category_scores}
\clearpage

\section{Category-specific EHQ1--EHQ2 overlap}

\begin{table}[htbp]
\centering
\caption{Exploratory category-specific association between EHQ1 and EHQ2 in
the fourteen-model confirmatory panel. Gaps are model means of
$EHQ_2-EHQ_1=CC/n$. Confidence intervals are percentile bootstrap intervals
using the frozen analysis settings (seed 42; 10,000 resamples).}
\label{tab:s-category-component-overlap}
\small
\begin{tabular}{lccccc}
\toprule
Category & Mean gap & Pearson $r$ & 95\% CI & Spearman $\rho$ & 95\% CI \\
\midrule
FEQ & $0.0000$ & $1.000$ & Identity & $1.000$ & Identity \\
CCQ & $0.0000$ & $1.000$ & Identity & $1.000$ & Identity \\
PCQ & $0.0479$ & $0.996$ & $[0.993,0.999]$ & $0.991$ & $[0.929,1.000]$ \\
HNQ & $0.2359$ & $0.910$ & $[0.588,0.972]$ & $0.669$ & $[0.122,0.933]$ \\
\bottomrule
\end{tabular}
\end{table}

For FEQ and CCQ, the correlations are identities induced by the absence of a
scorable correct substantive response; they are not statistical evidence of
construct equivalence. PCQ technically permits the scores to differ but offers
little separation in this panel. HNQ supplies the largest mean gap and rank
divergence, showing that the criteria can separate when correct substantive
answers occur often enough, while the wide intervals preclude a claim of
independent dimensions. With only fourteen model-level observations, these
percentile-bootstrap intervals are descriptive. As a small-sample sensitivity,
the Fisher-$z$ intervals are $[-.240,.734]$ for EHQ1--EHQ3,
$[-.142,.777]$ for EHQ2--EHQ3, and $[.969,.997]$ for EHQ1--EHQ2. They do not
change the interpretation in the main text.

\clearpage

\section{Response-classifier validation and score-impact sensitivity}

The validation sample contains 560 retained responses, with ten responses from
each of the fourteen analysed models in each category. The two coders completed
their files independently and without AI or external factual sources. Their
initial labels were frozen before disagreements were inspected. The 25
disagreements were subsequently resolved in a corresponding-author review with
disclosed AI decision support. Accordingly, the human--human statistics below
describe independent coding, while the all-item classifier comparison uses the
human-reviewed, AI-assisted adjudicated reference.

\ehqvalidationtable{agreement_summary}
\ehqvalidationtable{per_class_metrics}

The following post-hoc sensitivity recomputes EHQ1 and EHQ2 within the
validation sample from the automated and resolved human labels. Each model has
40 sampled responses. These quantities assess score direction and ordering in
the audit sample; they are not substituted for the full-run estimates and are
not interpreted as a correction model.

\ehqvalidationtable{score_impact_summary}
\ehqvalidationtable{score_impact_by_model}
\ehqvalidationtable{score_impact_bias}

\clearpage

\section{Category-balanced EHQ3 sensitivity}

Official EHQ3 pools substantive answers and therefore weights categories by the
number of such answers produced by each model. The following post-hoc analysis
first computes EHQ3 within each category and then averages the four category
values equally. It is a composition sensitivity, not a replacement score.

\ehqbalancedtable{category_balanced_ehq3_summary}
\ehqbalancedtable{category_balanced_ehq3_by_model}

\clearpage

\section{Matched CCQ selectivity anchor}

The capability probe restores the withheld span in 259 CCQ documents. The
table pairs each original missing-span item with its restored version. Because
the missing-span side uses the deterministic restraint label, this remains
sensitive to classifier error and does not substitute for matched answerable
controls in FEQ, PCQ, or HNQ.

\ehqselectivitytable{ccq_matched_selectivity}

\clearpage

\section{Additional measurement-audit diagnostics}

The first table supplies the per-model values underlying the ranges reported
in Section~6.6 of the main paper. The remaining tables quantify the post-hoc
family-associated classifier contrast, the corresponding human-reference EHQ1
ranks, contamination of the automated substantive-answer pool, its model-level
distribution, and the human-reference composite sensitivity. None is a
correction to the sealed full-run score.

\ehqtable{calibration_definition_sensitivity}
\ehqfamilytable{classifier_family_bias}
\ehqfamilytable{classifier_human_reference_ranks}
\ehqfamilytable{ehq3_pool_label_sensitivity}
\ehqcompositetable{ehq3_contamination_by_model}
\ehqcompositetable{classifier_composite_by_model}

\clearpage

\section{Alternative display of the overall ranking}
\begin{figure}[htbp]
\centering
\includegraphics[width=0.9\textwidth]{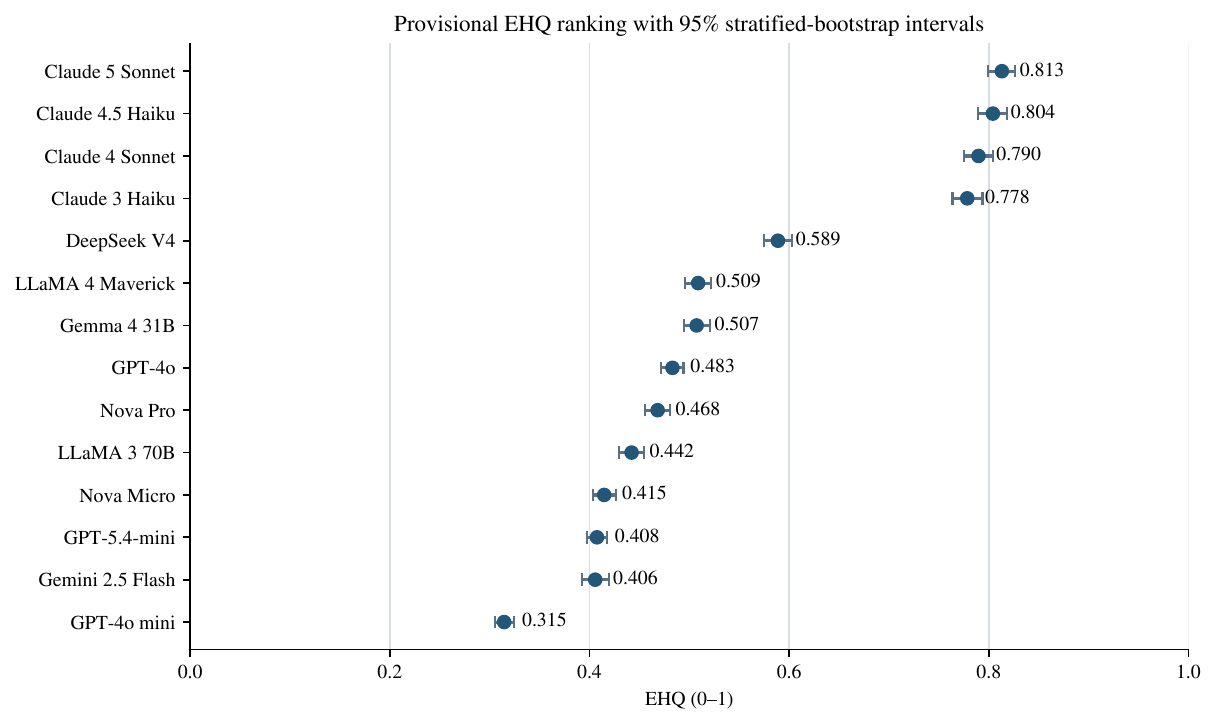}
\caption{EHQ for the fourteen-model confirmatory panel with 95\% stratified
bootstrap intervals. This forest display is supplementary because the main
text already reports exact scores and component profiles.}
\label{fig:s-ehq-ranking}
\end{figure}